\documentclass[10pt,journal,compsoc]{IEEEtran}
\ifCLASSOPTIONcompsoc
\usepackage[nocompress]{cite}
\else
\usepackage{cite}
\fi
\ifCLASSINFOpdf
\usepackage[pdftex]{graphicx}
\else
\fi
\usepackage{amsmath}
\usepackage{algorithmic}

\usepackage{array}
	\usepackage{bm}
	\usepackage{multicol}
	\usepackage{multirow}
	\usepackage{amssymb}
	\usepackage{booktabs}
	\usepackage[capitalize]{cleveref}
	\crefname{section}{Sec.}{Secs.}
	\Crefname{section}{Section}{Sections}
	\Crefname{table}{Table}{Tables}
	\usepackage{graphics}
	
	\usepackage{color}
	
	\usepackage{enumitem}
	
\begin{document}
		%
		\title{P$^2$SDF for Neural Indoor Scene Reconstruction}
		%
		%
		%
		%
		
		\author{Jing Li, Jinpeng Yu, Ruoyu Wang, Zhengxin Li, Zhengyu Zhang, Lina Cao, and Shenghua Gao 
			\IEEEcompsocitemizethanks{
				\IEEEcompsocthanksitem Jing Li and Jinpeng Yu contribute equally to this work;
				\IEEEcompsocthanksitem Corresponding Author: Shenghua Gao;
				\IEEEcompsocthanksitem Email: gaoshh@shanghaitech.edu.cn;
				\IEEEcompsocthanksitem Jing Li, Jinpeng Yu, Ruoyu Wang, Zhengxin Li and Shenghua Gao are with School of Information Science and Technology, ShanghaiTech University, Shanghai 201210, China;
				\IEEEcompsocthanksitem Jing Li is also with Shanghai Institute of Microsystem and Information Technology, Chinese Academy of Sciences, Shanghai 200050, China, and also with University of Chinese Academy of Sciences, China.
				\IEEEcompsocthanksitem Zhengyu Zhang is with The 38th Research Institute of China Electronic Technology Group Corporation, China.
				\IEEEcompsocthanksitem Lina Cao is with China Telecom Cloud Company, China.
				\IEEEcompsocthanksitem Shenghua Gao is also with Shanghai Engineering Research Center of Intelligent Vision and Imaging, and Shanghai Engineering Research Center of Energy Efficient and Custom AI IC, Shanghai 201210, China.}}
		
		%
		%

	\markboth{Journal of \LaTeX\ Class Files,~Vol.~14, No.~8, August~2015}%
	{Shell \MakeLowercase{\textit{et al.}}: Bare Demo of IEEEtran.cls for Computer Society Journals}
	%



	\IEEEtitleabstractindextext{%
		\begin{abstract}
			Given only a set of images, neural implicit surface representation has shown its capability in 3D surface reconstruction. However, as the nature of per-scene optimization is based on the volumetric rendering of color, previous neural implicit surface reconstruction methods usually fail in low-textured regions, including the floors, walls, etc., which commonly exist for indoor scenes. Being aware of the fact that these low-textured regions usually correspond to planes, without introducing additional ground-truth supervisory signals or making additional assumptions about the room layout, we propose to leverage a novel Pseudo Plane-regularized Signed Distance Field (P$^2$SDF) for indoor scene reconstruction. Specifically, we consider adjacent pixels with similar colors to be on the same pseudo planes. The plane parameters are then estimated on the fly during training by an efficient and effective two-step scheme. Then the signed distances of the points on the planes are regularized by the estimated plane parameters in the training phase. As the unsupervised plane segments are usually noisy and inaccurate, we propose to assign different weights to the sampled points on the plane in plane estimation as well as the regularization loss. The weights come by fusing the plane segments from different views. As the sampled rays in the planar regions are redundant, leading to inefficient training, we further propose a keypoint-guided rays sampling strategy that attends to the informative textured regions with large color variations, and the implicit network gets a better reconstruction, compared with the original uniform ray sampling strategy. Experiments show that our P$^2$SDF achieves competitive reconstruction performance in Manhattan scenes. Further, as we do not introduce any additional room layout assumption, our P$^2$SDF generalizes well to the reconstruction of non-Manhattan scenes. 
		\end{abstract}
		
		\begin{IEEEkeywords}
			Scene Reconstruction, Neural Surface Reconstruction, Implicit Representation, Plane-regularized Reconstruction.
	\end{IEEEkeywords}}

	\maketitle

	\IEEEdisplaynontitleabstractindextext

	%
	\IEEEpeerreviewmaketitle

	\IEEEraisesectionheading{\section{Introduction}\label{sec:introduction}}
	\IEEEPARstart{R}econstructing the 3D geometry of an indoor scene is an appealing and valuable task in computer vision because of its potential applications in computer graphics, virtual reality, etc. As the acquisition of 3D ground truth is expensive, it is promising to reconstruct the 3D scene directly from multiple images without 3D ground truth.
	Recently, neural radiance field (NeRF) \cite{mildenhall2021nerf} has achieved promising results in novel view synthesis based on implicit neural representations and volumetric rendering, giving a set of images. But NeRF does not explicitly regularize the geometry of the scene, resulting in poor reconstruction. To achieve a better geometry estimation, recent methods \cite{oechsle2021unisurf,wang2021neus,yariv2021volume} propose to model the density from the signed distance field instead of directly using the volume density in NeRF. However, indoor scenes have many textureless areas and repetitive patterns. It is difficult for previous methods to optimize the implicit network based on the rendering loss for these regions. As a result, the surface reconstruction of these methods at those planar regions is usually bumpy and noisy. It is necessary to incorporate additional constraints to get better reconstruction results.
	
	To improve the reconstruction for low-textured regions for indoor scenes, Manhattan-SDF \cite{guo2022neural} proposes to use a pre-trained semantic segmentation network to find the floor and wall regions, then regularize the geometry of these regions. The normal directions of floors and walls are regularized to be aligned with three dominant directions under Manhattan world assumption \cite{coughlan1999manhattan}. However, there are still many scenes with more than three dominant directions, which are the non-Manhattan scenes. Manhattan-SDF is not applicable to the non-Manhattan rooms. It may also perform poorly on scenes cluttered by objects with small visible areas of floors and walls, and its result is restricted by the segmentation network whose performance may degrade on scenes out of the pre-trained domain. 
	\begin{figure}[htbp]
		\centering
		\includegraphics[width=0.8\linewidth]{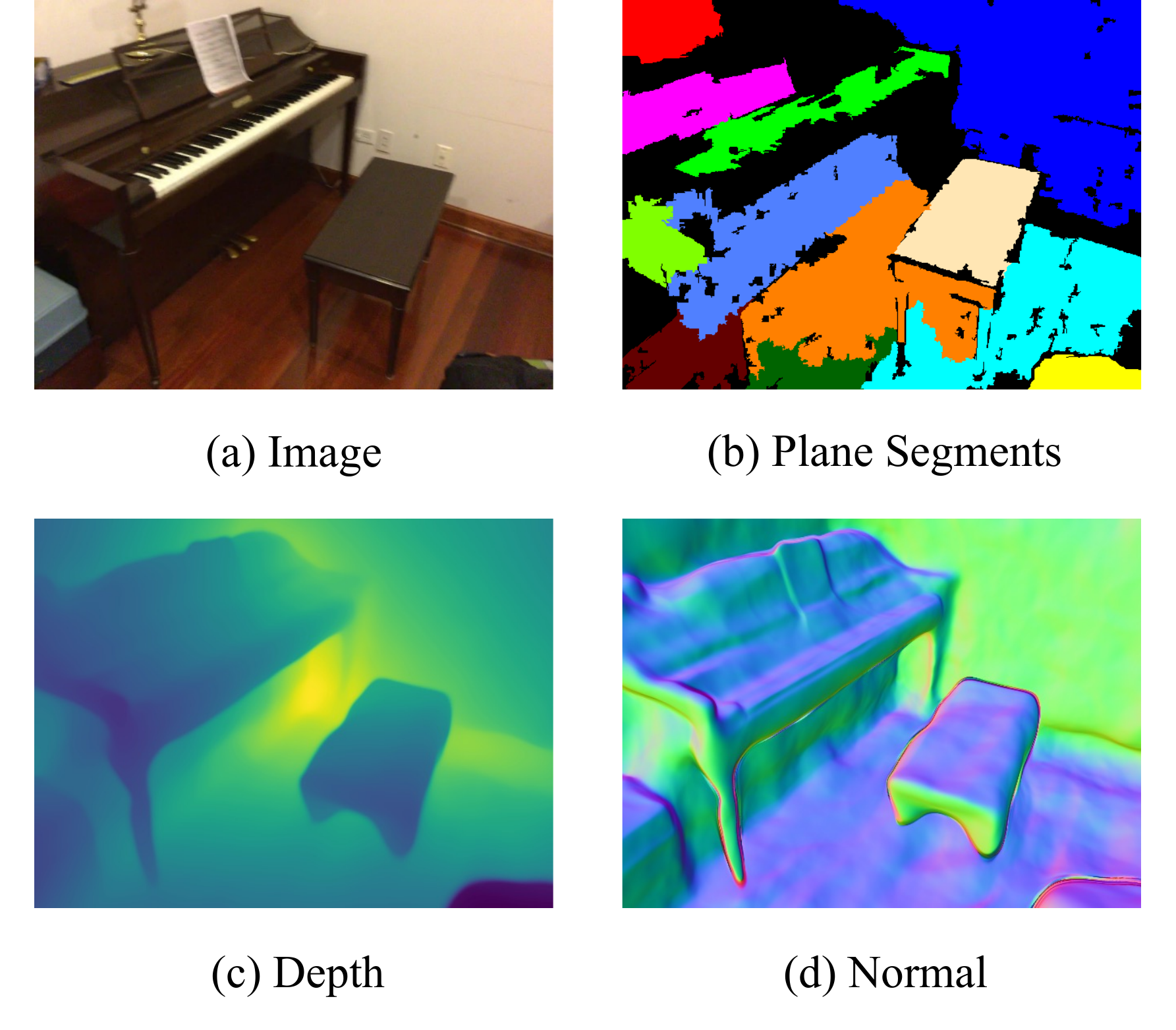}
		\caption{
			(a) is the image, (b) is the generated plane segments by super-pixel, (c) is the depth map, and (d) is the normal map from the reconstruction of P$^2$SDF. It can be seen that not only large areas of floor and wall regions but also small planar regions on objects, e.g., the chair and the piano, are segmented and reconstructed to be planar.
		}
		\label{fig:superpixel}
	\end{figure}
	In this work, we propose to tackle neural surface reconstruction of indoor scenes by using the pseudo planes as a regularization. To generate plane segments, we cluster the adjacent pixels with similar colors by super-pixels \cite{felzenszwalb2004efficient} and consider these regions as pseudo planes. The planes are not only large floor and wall regions but also small plane regions on objects, e.g., the chair and the piano exterior, as shown in \cref{fig:superpixel}. We then estimate the plane parameters on the fly during training in an effective and efficient two-step scheme and regularize the signed distance field (SDF) with these planes. As we do not make the Manhattan assumption on the indoor scene, our method generalizes well to both Manhattan and non-Manhattan rooms. 
	
	To be specific, we first sample only a few points on every plane and get their depth through the accumulated rendering. By back-projecting these points with the estimated depth, we fit the roughly estimated planes with these points by a least square method. The parameters of the roughly estimated plane are noisy and inaccurate as the number of sampled points is limited by the memory in the slow rendering. In the second step, we utilize the roughly estimated plane parameters to get the depth of a new set of sampled points and back-project them to the world space. The coordinates of these points are then fed into the MLPs to get the signed distances and the gradients of signed distances, i.e., normal directions. 
	We assume that the back-projected points on the roughly estimated planes are close enough to the planar surface, therefore the signed distances are also the distance to these planes. 
	We march these points according to the signed distances and the normal directions to get the coordinates of the points on the rectified planes, and the parameters of rectified plane are estimated with these points. In this step, we avoid using the depth with the expensive accumulated rendering and directly get the signed distances by querying points to get the rectified coordinates. Compared with the first step, this process is much faster, therefore the number of sampled points is much larger than that in the first step. After getting the rectified plane parameters, we regularize the signed distances of the newly sampled points on the roughly estimated planes by their distances to the rectified planes.
	
	As the unsupervised plane segments are often noisy and inaccurate, we propose to assign different weights to the points during the plane fitting and the plane regularization loss. The weights are obtained by fusing plane segments from different views. We conduct the fusion by predicting the segmentation probabilities and use the probabilities as the weights. The segmentation prediction is supervised by the super-pixel segments from different views, therefore considering the cross-view consistency implicitly. Existing MLP in VolSDF \cite{yariv2021volume} is queried by 3D coordinates to get the density and color. We introduce a new segmentation network branch to the MLP. The branch is supervised in the volume rendering by matching the accumulated prediction of the network, and the generated plane segments following \cite{carion2020end,wang2022dm}.
	During the fusion process, the segmentation network predicts lower probabilities for those inconsistent areas among different views, and higher probabilities for those consistent areas among different views. We utilize the probabilities as the weights assigned to different points during plane estimation and regularization.
	
	Further, we observe that the random ray sampling strategy which samples rays in volumetric rendering randomly is inefficient for indoor scenes. The reason is that the planar regions usually occupy large areas and the rays in the texture-less indoor planar regions are easy to be sampled in random sampling. Their colors are similar in the planar regions, therefore the RGB rendering loss for the sampled ray in the planar region is usually close to $0$ when there is still a large reconstruction error and contribute less to the optimization. To avoid such redundant sampling, we suppose that the ray sampling strategy should attend to the more informative textured regions with large color variations. We use Direct Sparse Odometry \cite{engel2017direct} to extract keypoints for the color image. We then treat the keypoints and the pixels around them as the informative regions and increase the sampling probability of the rays in these regions, therefore make them easier to be sampled. By our proposed keypoint guided rays sampling strategy, the implicit network gets a better reconstruction.

	The main contributions of this paper can be summarized as follows. i) We propose to constrain indoor neural surface reconstruction with pseudo planes, without imposing any assumption on the room layout or introducing additional supervision from external annotation on geometry. Experiments show that our method achieves competitive performance on Manhattan scenes and generalizes well to non-Manhattan scenes. ii) Under the memory and time restriction, we design an effective and efficient two-step strategy to estimate plane parameters, which are then used to regularize the signed distance field. iii) We introduce the weighting schedule on points during plane estimation and regularization. The weights are obtained by fusing the plane segments among views through volumetric rendering. iv) To avoid redundant rays sampling in the planar regions, we propose the keypoint guided sampling strategy to sample rays more efficiently and get a better performance. In this ray sampling strategy, the sampling probability in the informative textured regions is increased, therefore the rays are easier to be sampled.
	
	The rest of this paper is organized as follows: In \cref{sec:related_work}, we introduce the work related to the multi-view stereo and the neural scene reconstruction. In \cref{sec:method}, we detail our proposed P$^2$SDF, including pseudo plane regularization, plane segments fusion based points weighting, and keypoint guided ray sampling strategy. In \cref{sec:exp}, extensive experiments are conducted to validate the effectiveness and generalization of our method on the Manhattan scenes and the non-Manhattan scenes. We conclude our work in \cref{sec:conclusion}.
	
	\section{Related work}
	\label{sec:related_work}
	
	\noindent\textbf{Multi-view stereo.} Traditional 3D reconstruction methods from multi-view images \cite{bleyer2011patchmatch, merrell2007real, schonberger2016pixelwise, newcombe2011kinectfusion, schonberger2016structure, shotton2013scene} first estimate depth map for each view and then fuse the predicted depth from different views to obtain the final reconstruction result. However, these methods struggle with texture-less areas such as floor and wall regions in indoor scenes due to the mismatching on these regions by the hand-crafted features. To tackle this problem, some depth-based methods \cite{im2019dpsnet,yao2018mvsnet,yao2019recurrent, cheng2020deep} incorporate deep learning to extract global features, which is more robust for matching. After extracting global features, cost volumes are built and then fed to 3D CNNs to get final depth maps, and then leverage extra fusion process to reconstruct the scene. Others \cite{yin2019enforcing,kusupati2020normal} enforce the depth-normal consistency in the training process. Such methods usually reconstruct noisy and incomplete surfaces because of the inconsistency of individual predicted depth maps. Many methods also incorporate planar prior in MVS to achieve better performance on the texture-less regions. \cite{woodford2009global, gennert1988brightness} utilize the second-order priors on the smoothness of 3D surfaces as planar priors to constrain the planar areas in the scene. However, it leads to a triple cliques inference, making the optimization very difficult. \cite{gallup2010piecewise} chooses to train a planar classifier to distinguish the planar and non-planar areas, then links and fuses plane hypotheses across multiple overlapping views. The non-planar regions are reconstructed with the results of a standard multi-view stereo algorithm. Besides, some methods \cite{geiger2010efficient, xu2020planar} leverage the triangulation to build piece-wise planar regions. \cite{geiger2010efficient} choose to embed the priors into multi-view stereo directly over the disparity space. \cite{xu2020planar} uses the planar priors with PatchMatch MVS to focus on the geometric constraints of different views. Although PatchMatch-based MVS methods have great performance in scene reconstruction tasks, \cite{sun2021phi, romanoni2019tapa} point that PatchMatch-based methods often failed to reconstruct texture-less planar regions because the similarity measurement methods may be ineffective in these regions. \cite{sun2021phi} first generates the multiple plane hypotheses using the filtered depth maps on the areas which are not successfully recovered, then uses the Markov Random Field to select the depth hypotheses to improve the completeness of reconstruction results. \cite{romanoni2019tapa} proposes a novel PatchMatch hypothesis to improve the reconstruction completeness by expanding reliable depth estimation in neighboring texture-less regions. Inspired by these methods which prove the effectiveness of the planar constraints, our methods also utilize planar prior but in the implicit surface reconstruction process.
	
	\noindent\textbf{Neural scene reconstruction.} Different from the two-stage MVS methods, Atlas \cite{murez2020atlas} first proposed to end-to-end reconstruct the 3D scenes, where 2D CNN features are back-projected and accumulated into the voxel volume and predict the truncated signed distance function (TSDF) with 3D CNN. NeuralRecon \cite{sun2021neuralrecon} uses recurrent networks to fuse the features from different views, which is much faster. TransformerFusion \cite{bozic2021transformerfusion} uses the attention mechanism to attend to the most informative features of the image views in the temporal feature fusion. Recently, \cite{mescheder2019occupancy, mildenhall2021nerf, yariv2021volume, oechsle2021unisurf, wang2021neus, park2019deepsdf, sitzmann2019scene, wei2021nerfingmvs} represent the scene directly as an implicit neural function and generate high-resolution reconstruction result with low memory consumption. \cite{liu2020dist,niemeyer2020differentiable} propose a new method which can learning 3D shapes from only 2D images by introducing an implicit differentiable renderer. IDR \cite{yariv2020multiview} models the geometry with a zero level-set of a neural network and models the view-dependent appearance, but needs additional mask information. NeRF \cite{mildenhall2021nerf} achieves excellent novel view synthesis performance by implicit volume rendering but cannot get accurate and clean geometry. VolSDF \cite{yariv2021volume} and NeuS \cite{wang2021neus} derivate density from SDF in volume rendering, therefore achieving good reconstruction for objects. When reconstructing indoor scenes, these methods suffer from texture-less planar regions. To deal with the problem,  \cite{deng2022depth,roessle2022dense,lin2021efficient,wei2021nerfingmvs} utilizes the sparse depth from COLMAP \cite{schonberger2016structure} as additional supervision or to guide the sampling process. Manhattan-SDF \cite{guo2022neural} regularizes the normal directions of pixels in the floor and wall regions under the Manhattan world assumption \cite{coughlan1999manhattan}. The performance of Manhattan-SDF is limited by the pre-trained semantic segmentation network, which predicts the semantic floor and wall regions in indoor scenes. Moreover, it cannot generalize to non-Manhattan scenes and the scenes cluttered by objects with small visible areas of floors and walls. Our method constrains indoor scenes with pseudo planes obtained by unsupervised super-pixel segmentation. Therefore, our approach can handle the aforementioned complex scenes.
	\begin{figure*}[htbp]
		\centering
		\includegraphics[width=1.0\linewidth]{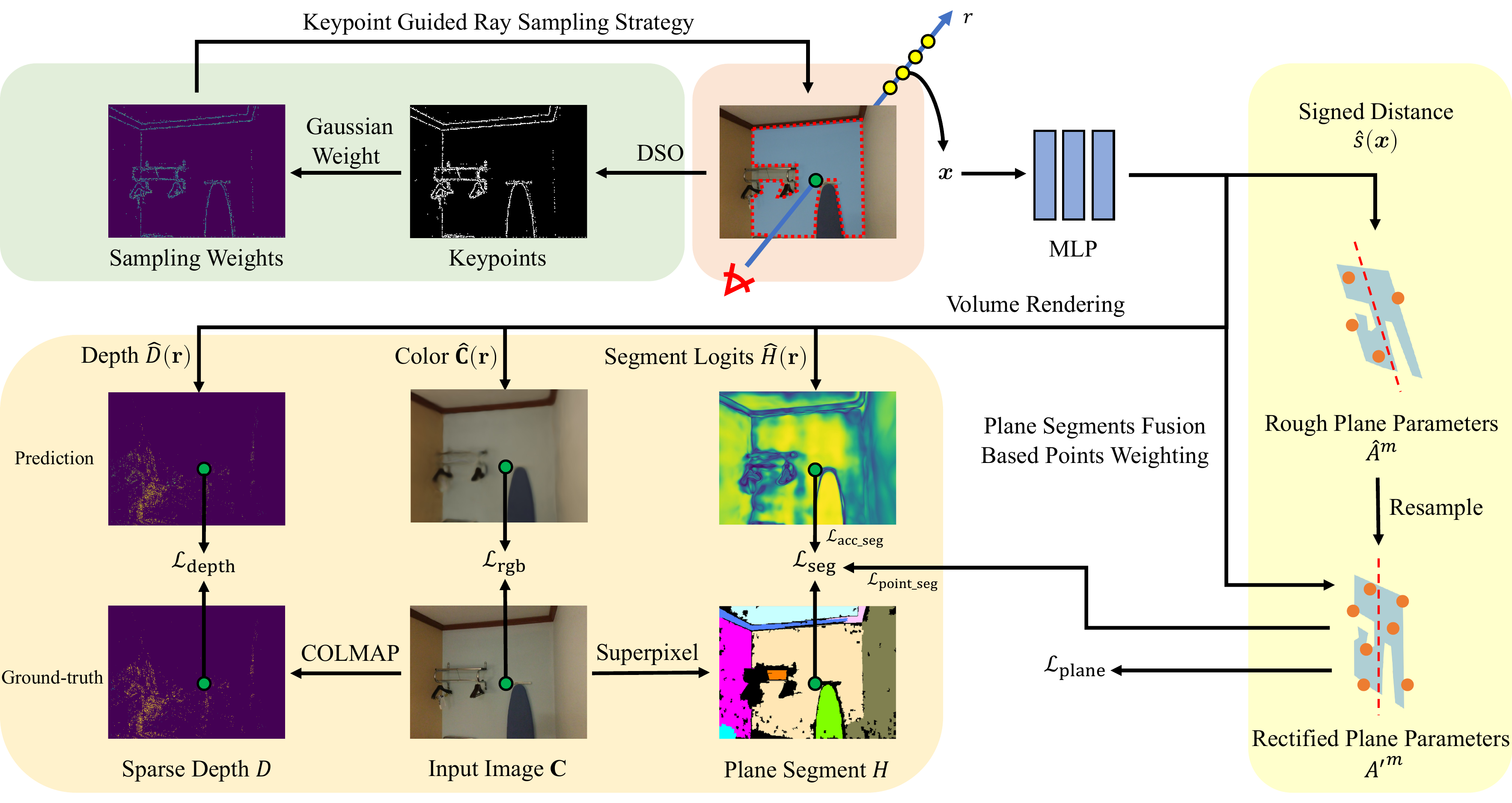}
		\caption{The overview of our method. We utilize the DSO method to extract keypoints and then calculate the ray sampling weight to guide the sampling. We learn the appearance, signed distance, depth, and segments of 3D scenes with implicit neural representations. For each ray, we use differentiable volume rendering to render its pixel color, depth, and segment probabilities, which are supervised with the input image, sparse depth obtained by COLMAP, and plane segments in 2D by super-pixel, respectively. We propose the novel plane regularization by a two-step plane parameter estimation strategy, which improves the smoothness in planar regions and details in the non-planar parts. We calculate rough plane parameters according to the accumulated depth of the sampled points and resample more points to compute the rectified plane parameters according to the signed distances and the normal directions. To overcome noisy and inaccurate plane segments, we additionally introduce plane segments fusion strategy to weight the points in the second rectified plane estimation step and the plane regularization.}
		\label{fig:overview}
	\end{figure*}
	
	\section{Method}
	\label{sec:method}
	We tackle indoor reconstruction through neural implicit surface reconstruction. However, the optimization of implicit surface reconstruction is difficult for the texture-less regions. Therefore, we propose pseudo plane regularization to apply additional regularization on the SDF of texture-less regions. As the generated plane segments are noisy and inaccurate, we propose to fuse the plane segments from different views and assign points with varied weights in plane estimation and regularization based on the fusion. Lots of rays are randomly sampled in the planar regions. Therefore, the RGB rendering loss for the sampled ray in the planar region is usually close to 0 with a large reconstruction error; therefore the random ray sampling in these regions is inefficient and contributes less to the optimization. To avoid such redundant sampling, we propose the keypoint guided ray sampling strategy to attend to the informative textured regions with large color variations. As a result, the implicit network gets a better reconstruction result. The overview of our method can be seen in \cref{fig:overview}.
	\subsection{Neural Implicit Surface Reconstruction}
	We reconstruct the scenes by the neural implicit surface reconstruction methods \cite{oechsle2021unisurf,wang2021neus,yariv2021volume} without 3D supervision. The reconstruction result is represented by a signed distance field (SDF). Following VolSDF \cite{yariv2021volume}, given one 3D point $\bm{x}$, we get its color $\mathbf{\hat{c}}(\bm{x})$\footnote{\^{} denotes the prediction from the network.} and signed distance $\hat{s}(\bm{x})$ by querying the MLP network $F$ as:
	\begin{equation}
		(\mathbf{\hat{c}}(\bm{x}), \hat{s}(\bm{x})) = F(\bm{x}).
	\end{equation}
	We derive the density $\sigma(\bm{x})$ from $\hat{s}(\bm{x})$ in volume rendering as:
	\begin{equation}
		\sigma(\bm{x};\beta)= \begin{cases}\frac{1}{2} \exp \left(\frac{\hat{s}(\bm{x})}{\beta}\right) & \text { if } \hat{s}(\bm{x}) \leq 0 \\ 1-\frac{1}{2} \exp \left(-\frac{\hat{s}(\bm{x})}{\beta}\right) & \text { if } \hat{s}(\bm{x})>0,\end{cases}
	\end{equation}
	where $\beta$ is a learnable parameter.
	
	The transparency $T$ of the $i$-th sampled point along the ray $\bm{r}$ in the discrete form is:
	\begin{equation}
		T_i = \exp(-\sum_{j=1}^{i-1} \delta_j\sigma_j),
	\end{equation}
	where $\delta_j=t_{j+1} - t_j$ is the distance between adjacent samples $t_{j}$ and $t_{j+1}$ along the ray.
	The accumulated color $\hat{\mathbf{C}}(\bm{r})$ along the ray $\bm{r}$ is:
	\begin{equation}
		\hat{\mathbf{C}}(\bm{r})=\sum_{i=1}^K T_i\left(1-\exp \left(-\sigma_i \delta_i\right)\right) \mathbf{\hat{c}}_i,
	\end{equation}
	where $K$ is the number of sampled points.
	The training loss contains three terms:
	\begin{equation}
		\mathcal{L} = \mathcal{L}_{\text{rgb}} + \lambda_{\text{sdf}}\mathcal{L}_{\text{sdf}} + \lambda_{\text{depth}}\mathcal{L}_{\text{depth}},
	\end{equation}
	where
	\begin{equation}
		\mathcal{L}_{\text{rgb}}=\sum_{\bm{r} \in \mathcal{R}}\|\hat{\mathbf{C}}(\bm{r})-\mathbf{C}(\bm{r})\|,
	\end{equation}
	\begin{equation}
		\mathcal{L}_{\text{sdf}}=\sum_{\bm{x} \in \mathcal{X}}\left(\left\|\nabla_{\bm{x}} d(\bm{x})\right\|_2-1\right)^2,
	\end{equation}
	$\mathbf{C}(\bm{r})$ is the ground-truth color of the ray $\bm{r}$. $\mathcal{L}_{\text{sdf}}$ is the Ekinoal loss \cite{yariv2020multiview,yariv2021volume} to guarantee the property of SDF. $\mathcal{X}$ is the set of points sampled uniformly in the space. $\mathcal{L}_{\text{depth}}$ is the depth loss supervised by the depth $D(\bm{r})$ from SfM \cite{schonberger2016structure} following \cite{deng2022depth,roessle2022dense,guo2022neural} to regularize the rendered depth $\hat{D}(\bm{r})$ as:
	\begin{equation}
		\hat{D}(\bm{r})=\sum_{i=1}^K T_i\left(1-\exp \left(-\sigma_i \delta_i\right)\right) t_i,
		\label{eq:depth}
	\end{equation}
	\begin{equation}
		\mathcal{L}_\text{depth}=\sum_{\bm{r} \in \mathcal{D}}|\hat{D}(\bm{r})-D(\bm{r})|,
	\end{equation}
	where $\mathcal{D}$ is the set of rays that have valid depth values estimated by SfM\cite{schonberger2016structure} on the image. Since the depth from SfM is sparse and is not available in the texture-less regions, the reconstruction result is still poor with the loss above. Therefore, it is necessary to incorporate stronger regularization to constrain the indoor scene reconstruction to achieve a better result.
	
	\subsection{Plane Regularization}
	There are many man-made planes in indoor scenes, not only walls, floors, and ceilings but also tables, wardrobes, etc. These areas are also texture-less areas without valid supervision in reconstruction. It is reasonable to improve reconstruction in these areas by regularizing these areas to be flat if we know that these areas are planes.
	
	\subsubsection{Plane Segments Generation}
	Plane segments can be obtained by plane detection \cite{liu2018planenet,liu2019planercnn,yu2019single}. However, plane detection methods need extra annotated planes, which are not easy to obtain. Therefore, we propose to get planes unsupervisedly by super-pixel segmentation. Given one image during training, we follow \cite{concha2014using,concha2015dpptam,yu2020p,li2021structdepth} to get the plane segments by super-pixel segmentation without supervision. The Felzenszwalb super-pixel segmentation\cite{felzenszwalb2004efficient} uses the greedy approach and segments areas with low gradients. The pixels with similar colors are clustered together. We treat the super-pixel segments whose areas are larger than the predefiend thresholds to be the pseudo planes, filtering the small segments which are likely to be non-planar out. As shown in \cref{fig:superpixel}, areas such as floors, walls, chair surface, and piano exterior are segmented.
	
	\begin{figure}[htbp]
		\centering
		\includegraphics[width=0.8\linewidth]{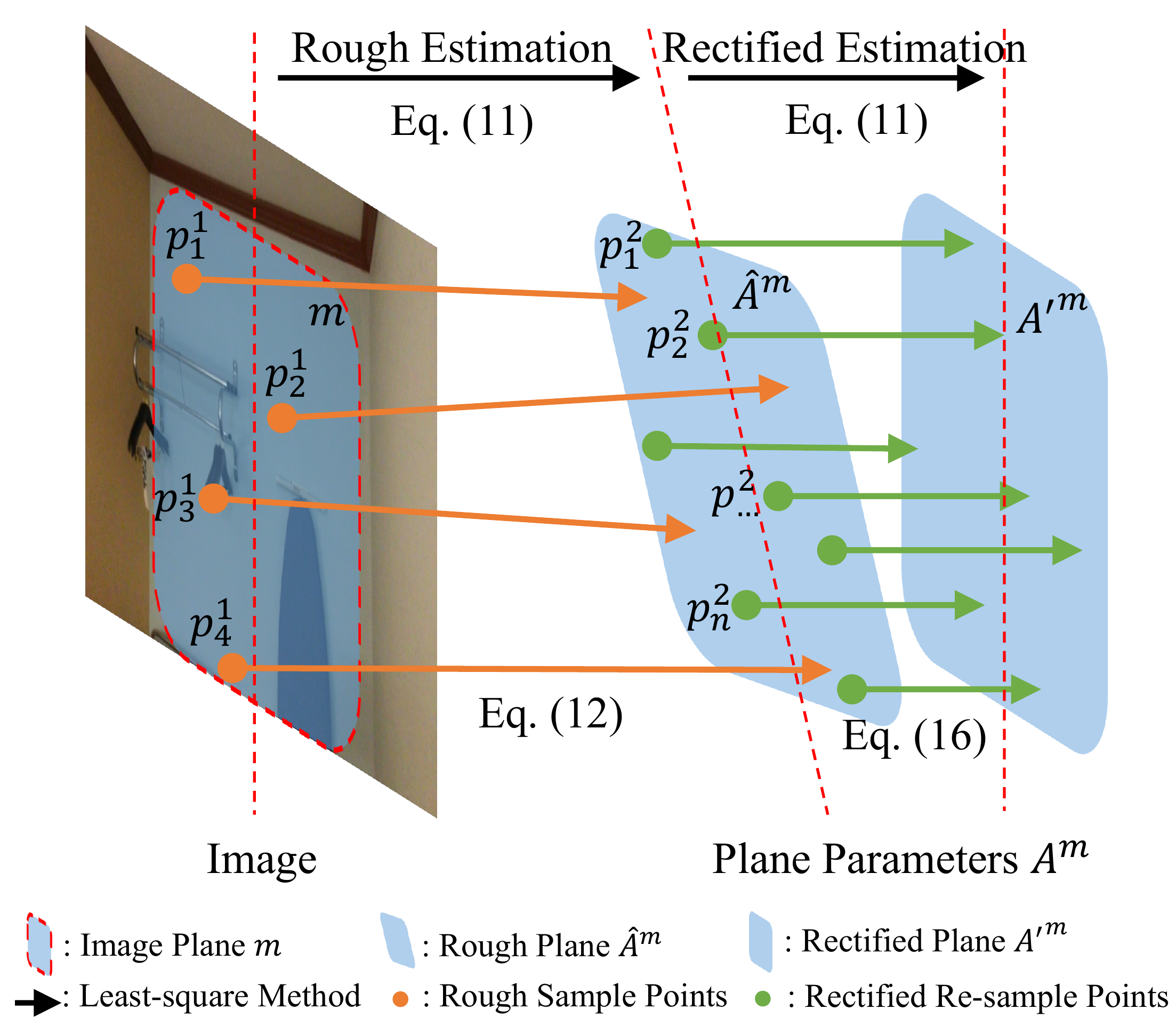}
		\caption{The illustration of the two-step plane parameter estimation. We first estimate the parameters of the rough planes by fitting a few points back-projected by the expensive rendering depth in the first step. Further, we randomly sample more points on the rough plane and then obtain the signed distances and the normal directions by directly querying the MLPs efficiently. Finally, we march the points according to their signed distances and normal directions, and estimate the parameters of the rectified planes by fitting these points.}
		\label{fig:plane_parameter}
	\end{figure}
	
	\subsubsection{Two-step Plane Parameter Estimation}
	\label{sec:plane}
	We use the least square method to get the $m$-th plane parameters $A^m$ as \cite{yu2020p}. Given a set of 3D points $X=[\bm{x}_1, \bm{x}_2, \cdots \bm{x}_{n_1}]^\top$ belonging to the $m$-th plane in one image during training, the relation between the points $X$ and the plane parameters $A^m$ is
	\begin{equation}
		XA^m=\bm{1},
	\end{equation}
	Then $A^m$ can be computed with a closed-form solution:
	\begin{equation}
		A^m=\left(X^{\top} X+\epsilon I\right)^{-1} X^{\top} \bm{1},
		\label{eq:plane}
	\end{equation}
	where $I$ is the identity matrix.
	Enough points are beneficial to get accurate plane parameters. However, as we introduce plane regularization on the fly during training, we need to estimate the plane parameters online. To get the coordinates of 3D points, it is necessary to get the depth of these points, which is rendered by volume rendering as \cref{eq:depth}. It is slow and memory-consuming to query every point along the ray to get the depth of only one point. As a result, there are few points to estimate per plane's parameters. Therefore, we propose to get the plane parameters efficiently and effectively in a two-step scheme, utilizing the property of SDF. 
	
	Our two-step plane parameter estimation scheme includes the rough plane parameter estimation and the rectified plane parameter estimation, as can be seen in \cref{fig:plane_parameter}. 
	In the rough plane parameters estimation, we sample only a few points on every plane and get their depth through the accumulated rendering. By back-projecting these points with the estimated depth, we fit the roughly estimated planes. The parameters of the roughly estimated plane are noisy and inaccurate as the number of sampled points is limited by the memory in the slow rendering process. In the second step, we sample more points on the planes from the first step and directly feed these points to MLPs to get the signed distances and the normal directions. We then calculate the plane parameters accurately according to the signed distances and the normals. In this step, the expensive accumulation for depth is avoided. Therefore we can sample more points and the plane parameters are more accurate. We detail our two step scheme and the pseudo plane regularized SDF in the following.\\

	\noindent\textbf{Rough plane parameter estimation.} In the first step, we get the roughly estimated plane parameters, which is inaccurate but fast. We sample four points $\bm{p}^1_1\text{, }\bm{p}^1_2 \text{, } \bm{p}^1_3 \text{, } \bm{p}^1_4 \in \mathbb{R}^3$ in the pixel space\footnote{Here $\bm{p}^1_1\text{, }\bm{p}^1_2 \text{, } \bm{p}^1_3 \text{, } \bm{p}^1_4$ are the homogeneous coordinates of the sampled points. The superscript $^1$ and $^2$ are denoted to distinguish the points in the first step rough plane parameter estimation and the second step rectified plane parameter estimation, respectively} per plane segment during rendering and get the corresponding rendered depth $d^1_1\text{, }d^1_2\text{, }d^1_3\text{, } d^1_4$. The corresponding 3D points in the camera space of the training image are calculated as follows:
	\begin{equation}
		\bm{x}^1_i=d^1_i K^{-1} \bm{p}^1_i,
	\end{equation}
	where $K$ is the camera intrinsics matrix. We then get the roughly estimated plane parameters $\hat{A}^m$ based on \cref{eq:plane}.
	
	\noindent\textbf{Rectified plane parameter estimation.} In the second step, we re-estimate the plane parameters utilizing SDF. We re-sample a new set of points $\bm{p}^2_1\text{, }\bm{p}^2_2\text{,} \dots\text{,} \bm{p}^2_{n_2}$ on the roughly estimated plane in the pixel space. The depth of these points on the roughly estimated plane is calculated as follows:
	\begin{equation}
		d^2\left(\bm{p}^2_i\right)=\left(\hat{A}^{m^{\scriptstyle\top}} K^{-1} \bm{p}^2_i\right)^{-1}.
	\end{equation}
	We then back-propagate these points to get the coordinates of the 3D points in the world space as:
	\begin{align}
		\bm{x}^2_i &= E^{-1}d^2(\bm{p}^2_i)K^{-1}\bm{p}^2_i\\
		&=E^{-1}\bm{p}_i^{2^{\scriptstyle-1}}K{\hat{A}^{m^{\scriptstyle{-\!\!\top}}}}K^{-1}\bm{p}^2_i,
	\end{align}
	where $E$ is the extrinsic matrix of the camera. We then feed the 3D coordinates of these points into the MLP to get the corresponding signed distances $\hat{s}(\bm{x}^2_i)$ and normals $\hat{\bm{n}}(\bm{x}^2_i)$, respectively. The normal is computed by the derivative of the signed distance $s$. We assume that the points on the roughly estimated planes are close enough to the true surface, i.e., the plane. Therefore, we get the coordinates of the points on the rectified plane by marching these points according to the signed distances and normals as:
	\begin{equation}
		\bm{\bar{x}}_{i} = \bm{x}^2_i - \hat{s}(\bm{x}^2_i)\hat{\bm{n}}(\bm{x}^2_i)
	\end{equation}
	The rectified plane parameters ${A^\prime}^m$ is calculated from these points. Here the coordinates of the newly sampled points are directly from the MLP without the accumulated depth. Therefore, the number of sampled points in this step is much larger than in the first step. As a result of the scheme, the calculation of plane parameters is efficient and effective.
	\subsubsection{Pseudo Plane regularized SDF}
	After getting the rectified plane parameters ${A^\prime}^m$, we calculate the signed distances $s(\bm{x}^2_i)$ of the sampled points $\bm{x}^2_i$ to the estimated plane as:
	\begin{equation}
		s(\bm{x}^2_i) =\text{sgn}({\bm{\bar{x}}_i}, \bm{x}^2_i, \bm{x}_c)	\|\frac{{\bm{x}_i^{2^{\scriptstyle\top}}}{A^\prime}^m - 1}{\sqrt{A^{\prime^{m^{\scriptstyle\top}}} {A^\prime}^m}}\|,
	\end{equation}
	where
	\begin{equation}
		\text{sgn}({\bm{\bar{x}}_i}, \bm{x}^2_i, \bm{x}_c) = \begin{cases}1, & \text{if } \|{\bm{\bar{x}}}_i - \bm{x}_c\| \geq \|{\bm{x}}_i - \bm{x}_c\|, \\ -1, & \text{if } \|{\bm{\bar{x}}}_i - \bm{x}_c\| < \|{\bm{x}}_i - \bm{x}_c\|, \end{cases}
	\end{equation}

$\bm{x}_c$ is the coordinate of the camera, and $\text{sgn}(\cdot)$ is the sign function that turns positive if the sampled point is outside the surface and negative if inside. We consider the calculated signed distance as the pseudo `ground truth' to regularize the SDF. The plane regularization loss for the predicted signed distances $\hat{s}(\bm{x}_i)$ during training is:
\begin{equation}
	\mathcal{L}_{\text{plane}} = \sum_{\bm{x} \in {\mathcal{X}}^2}\|s(\bm{x}) - \hat{s}(\bm{x})\|.
	\label{eq:l_plane}
\end{equation}

\subsection{Plane Segments Fusion Based Points Weighting}
The plane segments are usually noisy and inaccurate. If the points outside the true plane are considered in the pseudo plane, the plane parameters is likely to be poorly estimated and the points are also forced to be close to the plane by the false regularization. Therefore, we propose to weight the points differently in plane estimation and regularization to remedy these problems. The weights are obtained by fusing plane segments from different views. We assume that the inconsistent areas of the segments from different views are more likely to be poorly segmented and the consistent areas are more accurate.

\noindent\textbf{Plane Segments Fusion.}
We propose and train a segmentation network branch attached to the MLPs to fuse the plane segments. The network predicts sigmoid probability $h^m(\bm{x})$ of point $\bm{x}$ on plane $m$. To get the segmentation prediction $\hat{H}^m$ of plane $m$ on the image, we accumulate the segmentation probability in volume rendering as:
\begin{equation}
	\hat{H}^m(\bm{r})=\sum_{i=1}^K T_i\left(1-\exp \left(-\sigma_i \delta_i\right)\right) h^m_i,
\end{equation}
We train the segmentation network branch with the binary cross entropy (BCE) loss:
\begin{equation}
	\notag\mathcal{L}_{\text{seg}} = \notag\mathcal{L}_{\text{acc\_seg}} + \notag\mathcal{L}_{\text{point\_seg}}.
\end{equation}
The segmentation loss contains two parts. $\notag\mathcal{L}_{\text{acc\_seg}}$ is the loss for the accumulated segmentation sigmoid probability, and $\notag\mathcal{L}_{\text{point\_seg}}$ is for the sigmoid probability $\mathbf{h}^i(\bm{r})$ of the sampled points $\bm{x}^2$ on the roughly estimated planes on the ray $\bm{r}$ in the second step.
\begin{align}
	\notag\mathcal{L}_{\text{acc\_seg}}=
	\sum_{\bm{r} \in \mathcal{R}} \sum_{i=1}^M H^i(\bm{r}) &\log(\hat{H}^{\alpha(i)}(\bm{r}))\\
	+ (1 - H^i(\bm{r})) & \log(1 - \hat{H}^{\alpha(i)}(\bm{r})),
\end{align}
\begin{align}
	\notag\mathcal{L}_{\text{point\_seg}}=
	\sum_{\bm{r} \in \mathcal{R}} \sum_{i=1}^M h^i(\bm{r}) &\log(\hat{H}^{\alpha(i)}(\bm{r}))\\
	+ (1 - h^i(\bm{r})) & \log(1 - \hat{H}^{\alpha(i)}(\bm{r})).
\end{align}
The corresponding $i$ and $\alpha(i)$ is achieved by matching $H^i$ and $\hat{H}^{\alpha(i)}$ with Hungarian matching \cite{kuhn1955hungarian,carion2020end,wang2022dm}. We search for a permutation of $M$ elements $\alpha \in \mathfrak{S}_M$ with the lowest cost as:
\begin{equation}
	\alpha=\underset{\alpha \in \mathfrak{S}_M}{\arg \min } \sum_i^M \mathcal{L}_{\text {match }}\left(H^i, \hat{H}^{\alpha(i)}\right)
\end{equation}
$\mathcal{L}_\text{match}$ is measured by pair-wise sIoU and BCE.

\noindent\textbf{Weighting.} By fusing the plane segments, $h^m(\bm{x})$ is relatively low in the inconsistent regions, while in the consistent regions, $h^m(\bm{x})$ is high. We normalize $h^m(\bm{x_i})$ to $\bar{h}^m(\bm{x_i}) = \frac{h^m(\bm{x_i})}{\sum_{i=1}^n h^m(\bm{x_i})}n$. We utilize $h^m(\bm{x})$ to weight \cref{eq:plane} in the second step rectified plane parameter estimation as:
\begin{align}
	W=\begin{bmatrix}
		\bar{h}^m(\bm{x_1}) & &\\& \ddots &\\& & \bar{h}^m(\bm{x_{n_2}})
	\end{bmatrix},\\
	A=\left(X^{\top}W X+\epsilon I\right)^{-1} X^{\top} Y.
\end{align}
\cref{eq:l_plane} in the plane regularization is weighted as:
\begin{equation}
	\mathcal{L}_{\text{plane}} = \sum_{\bm{x} \in {\tilde{\mathcal{X}}}^2} \bar{h}^m(\bm{x})\|s(\bm{x}) - \hat{s}(\bm{x})\|.
\end{equation}
Here we do not weight the points in first step plane parameters estimation. The reason is that we only get the roughly estimated plane parameters in the first step. The second step is more important.

\begin{table*}[htbp]
	\centering
	\small
	\renewcommand{\tabcolsep}{1.5ex}
	\caption{Results on the ScanNet dataset and the 7-Scenes dataset. We compare our method with the classical MVS and the neural volume rendering methods on the four indoor scenes from Manhattan-SDF \cite{guo2022neural} in each dataset. $^\dag$ indicates the result reproduced by ours. Other results are from Manhattan-SDF. The averaged results show that our method achieves the competitive reconstruction performance on the ScanNet dataset and performs better on the 7-Scenes scenes.}
	\begin{tabular}{l|ccccc|ccccc}
		\toprule
		\multirow{2}{*}{Method} & \multicolumn{5}{c|}{ScanNet} & \multicolumn{5}{c}{7-Scenes} \\
		\cmidrule{2-11}
		& Acc $\downarrow$ & Comp $\downarrow$ & Prec $\uparrow$ & Recall $\uparrow$ & \textbf{F-score} $\uparrow$ & Acc $\downarrow$ & Comp $\downarrow$ & Prec $\uparrow$ & Recall $\uparrow$ & \textbf{F-score} $\uparrow$ \\
		\midrule
		NeRF \cite{mildenhall2021nerf}    & 0.735 & 0.177 & 0.131 & 0.290 & 0.176 & 0.573 & 0.321 & 0.159 & 0.085 & 0.083 \\
		NeuS \cite{wang2021neus}          & 0.179 & 0.208 & 0.313 & 0.275 & 0.291 & 0.151 & 0.247 & 0.313 & 0.229 & 0.262 \\
		VolSDF \cite{yariv2021volume}     & 0.414 & 0.120 & 0.321 & 0.394 & 0.346 & 0.285  & \textbf{0.140} & 0.220 & \textbf{0.285} & 0.246 \\
		COLMAP \cite{schoenberger2016sfm} & \textbf{0.047} & 0.235 & \textbf{0.711} & 0.441 & 0.537 & \textbf{0.069} & 0.417 & \textbf{0.536} & 0.202 & 0.289 \\
		Manhattan-SDF$^\dag$ \cite{guo2022neural} & 0.050 & \textbf{0.055} & 0.705 & 0.660 & 0.681 & 0.161 & 0.288 & 0.360 & 0.182 & 0.239 \\
		\midrule
		Ours                 & 0.057 & 0.062 & 0.703 & \textbf{0.664} & \textbf{0.683} & 0.104 & 0.301 & 0.511 & 0.209 & \textbf{0.295} \\
		\bottomrule
	\end{tabular}
	\label{tab:ori_scenes}
\end{table*}

\subsection{Keypoint Guided Ray Sampling Strategy}
\begin{figure}[htbp]
	\centering
	\includegraphics[width=1.0\linewidth]{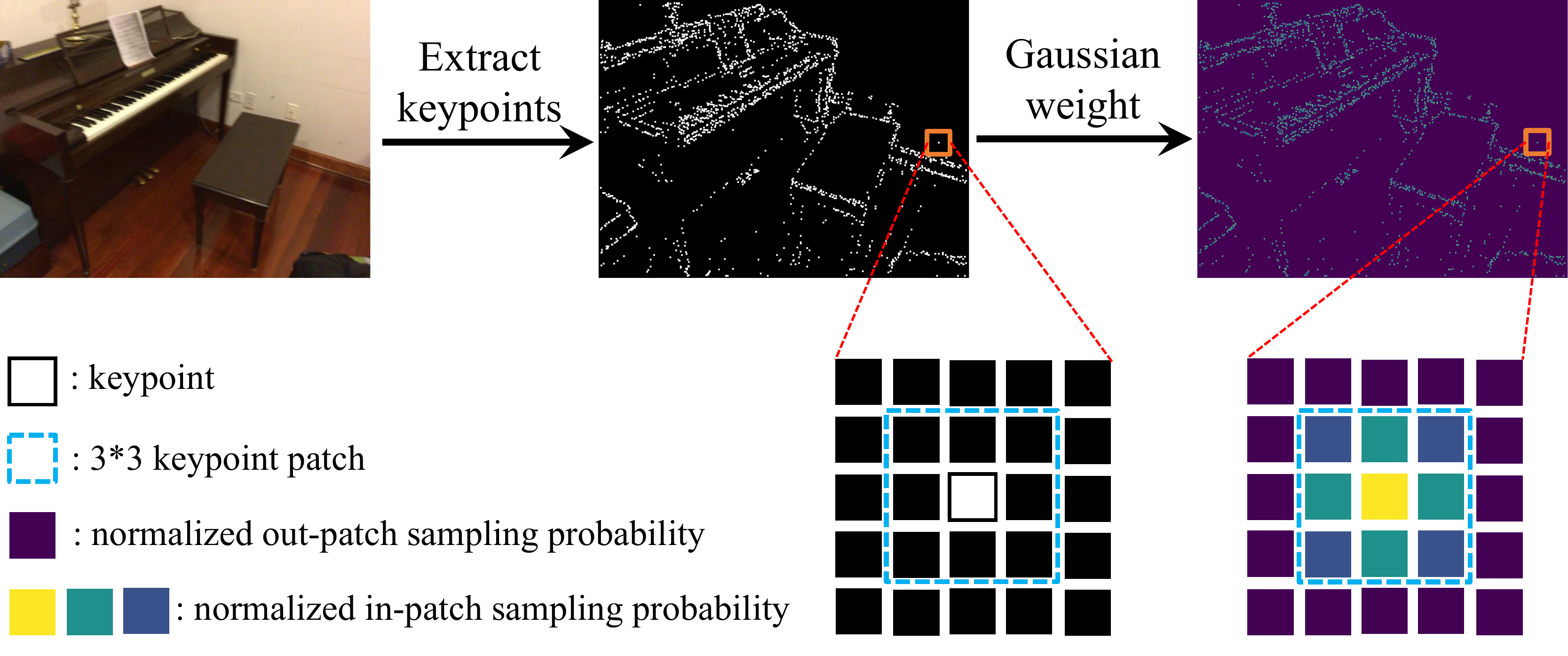}
	\caption{The illustration of keypoint guided ray sampling strategy. We utilize the Direct Sparse Odometry (DSO) \cite{engel2017direct} approach to extract keypoints from the color images. We initialize the sampling weight of every pixel in the image as 1, then choose the 3*3 patch centered at the keypoints and calculate the sampling weight of pixels in patches by Gaussian distance. We finally normalize the weights of the entire image to sample rays.}
	\label{fig:diagram_dso}
\end{figure}
Neural volumetric rendering is optimized based on the rendering loss between the predicted RGB color and the ground-truth color. In indoor scenes, ordinary random ray sampling is inefficient and redundant. The reason is that the planar regions usually occupy large areas and the rays in the texture-less regions are easy to be sampled. Considering there is a small offset for a sampled ray but still hitting inside the same plane. The RGB rendering loss is usually small as the RGB colors are similar inside the plane, after some iterations. Though the reconstruction is wrong, the loss may be close to zero.
Sampled rays in these regions contribute less to the reconstruction optimization. It is reasonable that the sampling strategy should attend to the informative textured areas with the large color variations. Inspired by \cite{furukawa2009accurate,schonberger2016pixelwise,schonberger2016structure,engel2017direct,yu2020p}, we propose the keypoint guided ray sampling strategy. To be specific, we extract the keypoints by Direct Sparse Odometry \cite{engel2017direct}. We calculate the sampling weight $G_{p_i}$ of 2D point $p_i$ by the Gaussian distance to the keypoints in the patch $P_{q_j}$ centered at the 2D keypoints $q_j$ as:
\begin{equation}
	G_{p_i} = \begin{cases} 1 + k\exp(\frac{{\|p_i - q_j\|}_2}{\gamma}), &j = \arg\min{\|p_i - q_j\|}_2, \\ &\text{and } p_i\in P_{q_j},\\ 1, &\text{otherwise}, \end{cases}
	\label{eq:sampl_weight}
\end{equation}
where $k$ and $\gamma$ are hyper-parameters to control the scale of sampling weights.
The final sampling probability $\bar{G}_{p_i}$ for point $p_i$ is normalized as:
\begin{equation}
	\bar{G}_{p_i} = \frac{G_{p_i}}{\sum_{j} G_{p_j}}
\end{equation}
The illustration of our proposed keypoint guided ray sampling strategy can be seen in \cref{fig:diagram_dso}.

\section{Experiments}
\label{sec:exp}

\subsection{Datasets and Metrics}
\noindent\textbf{Datasets.} We perform our experiments on ScanNet (V2) \cite{dai2017scannet} and 7-Scenes \cite{shotton2013scene}. ScanNet (V2) is a large-scale challenging real-world indoor scene dataset with ground-truth camera poses and surface reconstructions, containing around 2.5M RGB-D images captured in 1613 indoor scenes. 7-Scenes consists of RGB-D frames captured by a tracked Kinect RGB-D camera and uses KinectFusion \cite{izadi2011kinectfusion} to obtain dense 3D models and camera poses. 
While previous neural indoor scene reconstruction methods with depth and normal priors mainly focused on the scenes with Manhattan world assumption, in our experiments, we prove that our method can perform better in more general scenes. Thus we consider: a) The same real-world indoor scenes following \cite{guo2022neural}; b) Four additional Manhattan world scenes where the DeepLab V3+ segmentation network predicts poorly on the wall and floor regions; c) Four additional Manhattan world scenes with small visible areas of walls and floors; d) Four non-Manhattan world scenes from the ScanNet dataset.

\noindent\textbf{Metrics.} Following \cite{guo2022neural, sun2021neuralrecon, yu2022monosdf}, we use RGB-D fusion results as 3D reconstruction ground truth and evaluate our method using five standard metrics defined in \cite{murez2020atlas}: accuracy, completeness, precision, recall, and F-score with a threshold of 5cm.

\subsection{Implementation Details}
Our network is trained for 50 epochs with batches of 1024 rays. We use the Adam optimizer \cite{kingma2014adam} with the initial learning rate $5\times10^4$ and an exponential learning rate decay strategy. 
All RGB images are initially resized to $640\times480$ for scene reconstruction, and we utilize the Felzenszwalb super-pixel segmentation \cite{felzenszwalb2004efficient} and COLMAP \cite{schonberger2016structure} method to extract super-pixel and sparse depth maps from these re-scaled images respectively.
We implement our method with PyTorch \cite{paszke2019pytorch} and train our network on one NVIDIA GeForce RTX 2080Ti GPU. 
We obtain the camera intrinsic and extrinsic parameters 
from the original dataset \cite{dai2017scannet}, and adjust them to be consistent with the image scaling. Following \cite{atzmon2020sal}, we normalize all cameras to be inside a unit sphere and initialize network parameters by representing the initial SDF to a unit sphere with the surface normals facing inside. We evaluate the results by the Marching Cubes algorithm \cite{lorensen1987marching} to extract surface mesh from the learned signed distance function. To overcome the effect of unobserved regions in evaluation, suggested by Manhattan-SDF \cite{guo2022neural}, we render depth maps from predicted mesh and re-fuse them with TSDF fusion \cite{izadi2011kinectfusion} following \cite{murez2020atlas}.

\subsection{Comparisons with Other Methods}
\begin{figure*}[htbp]
	\centering
	\includegraphics[width=1.0\linewidth]{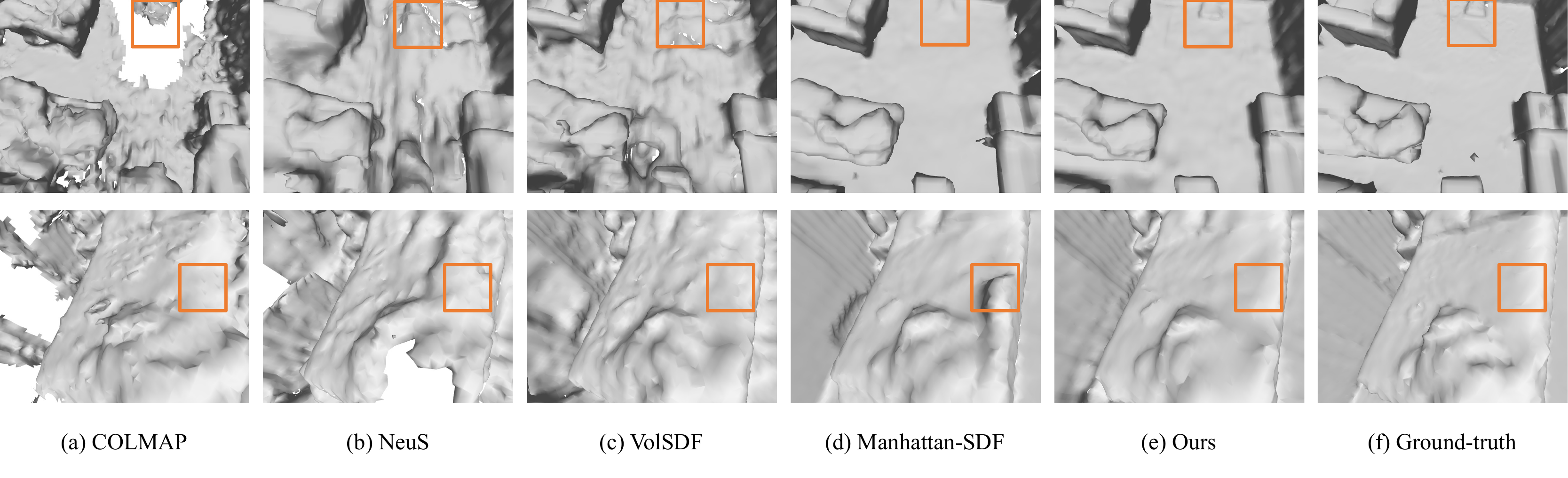}
	\caption{Visualization compared with other methods on the indoor scenes in Manhattan-SDF \cite{guo2022neural} from the ScanNet dataset. COLMAP \cite{schonberger2016structure} and NeuS \cite{wang2021neus} cannot produce proper reconstruction results. VolSDF \cite{yariv2021volume} leads to noisy reconstructions without additional constraints. Manhattan-SDF improves the smooth results in the regions such as floor, wall, and bed, but details are missing in some small planar areas such as the sweeping robot. In contrast, our method can get a more detailed surface in these regions, achieving compelling results.}
	\label{fig:ori_scenes}
\end{figure*}

We compare our method with the following methods: traditional MVS reconstruction method: COLMAP \cite{schonberger2016structure}. We use screened Poisson Surface reconstruction (sPSR) \cite{kazhdan2013screened} to reconstruct mesh from extracted point clouds; Neural volume rendering methods: NeRF \cite{mildenhall2021nerf}, NeuS \cite{wang2021neus} and VolSDF \cite{yariv2021volume}; Volume rendering with additional pre-trained segmentation model: Manhattan-SDF \cite{guo2022neural}. We do not visualize the reconstruction results of NeRF because it fails to reconstruct indoor scenes without geometric constraints. Our method achieves state-of-the-art performance not only in the Manhattan scenes but also in the non-Manhattan scenes. Our method is also unrestricted by the pre-trained network which may perform poorly on some scenes, and can also perform well on the cluttered indoor rooms with small visible areas of floors and walls.

\begin{table}[htbp]
	\centering
	\small
	\caption{Results compared with other methods on Manhattan scenes where DeepLab V3+ \cite{chen2018encoder} segmentation network predicts poorly on the wall and floor regions. We choose four scenes from the ScanNet dataset. Our approach can still perform well while Manhattan-SDF \cite{guo2022neural} degrades.}
	\begin{tabular}{l|ccccc}
		\toprule
		\makebox[0.08\linewidth][l]{Method} & \makebox[0.09\linewidth][c]{Acc $\downarrow$} & \makebox[0.09\linewidth][c]{Comp $\downarrow$} & \makebox[0.09\linewidth][c]{Prec $\uparrow$} & \makebox[0.09\linewidth][c]{Recall $\uparrow$} & \makebox[0.09\linewidth][c]{\textbf{F-score} $\uparrow$} \\
		\midrule
		NeRF \cite{mildenhall2021nerf}    & 0.420 & 0.612 & 0.041 & 0.017  & 0.023 \\
		NeuS \cite{wang2021neus}          & 0.283 & 0.482 & 0.182 & 0.136 & 0.153 \\
		VolSDF \cite{yariv2021volume}     & 0.254 &	0.306 &	0.239 &	0.179 &	0.198 \\
		COLMAP \cite{schoenberger2016sfm} & 0.131 &	\textbf{0.149} & 0.482 & 0.452 & 0.460 \\
            Manhattan-SDF \cite{guo2022neural} & \textbf{0.104} & 0.171 & 0.563 & 0.442 & 0.493 \\
		\midrule
		Ours                 & 0.107 & 0.168 & \textbf{0.563} & \textbf{0.472} & \textbf{0.512} \\
		\bottomrule
	\end{tabular}
	\label{tab:bad_iou}
\end{table}

\begin{figure*}[htbp]
	\centering
	\includegraphics[width=1.0\linewidth]{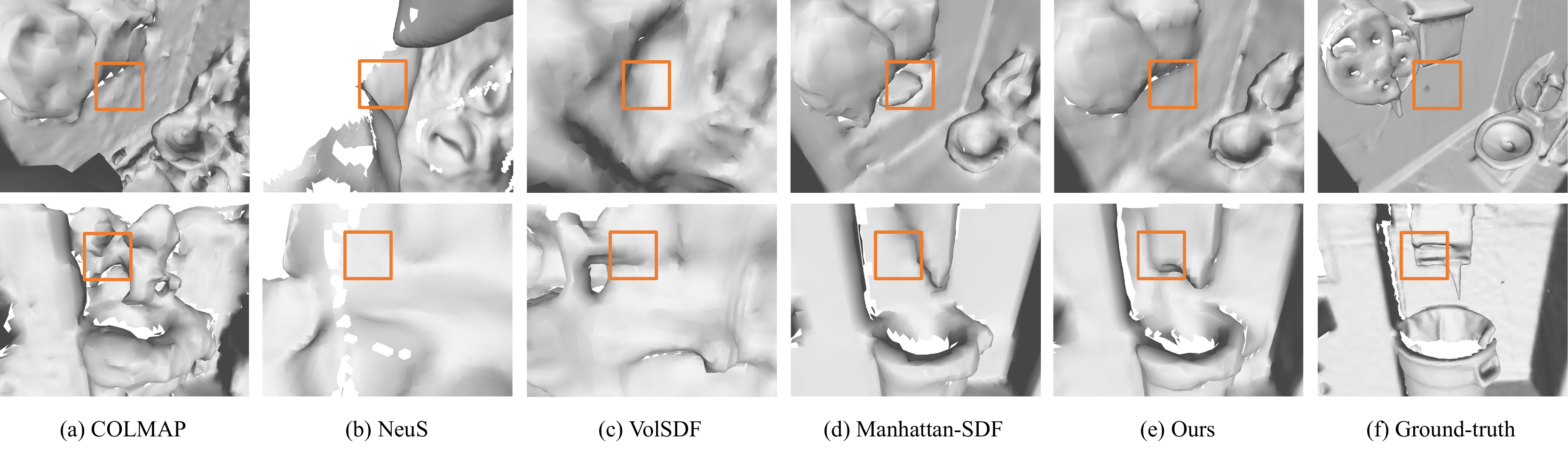}
	\caption{Visualization compared with other methods on Manhattan scenes where the DeepLab V3+ \cite{chen2018encoder} segmentation network predicts poorly on the wall and floor regions. COLMAP, NeuS, and VolSDF cannot achieve competitive results. Manhattan-SDF produces unreasonable reconstruction results in the poorly predicted segmentation regions, e.g., the wall region near the paper holder. Compared with these approaches, our approach can produce more details and smooth surfaces, achieving better results.}
	\label{fig:bad_iou}
\end{figure*}

\begin{figure*}[htbp]
	\centering
	\includegraphics[width=1.0\linewidth]{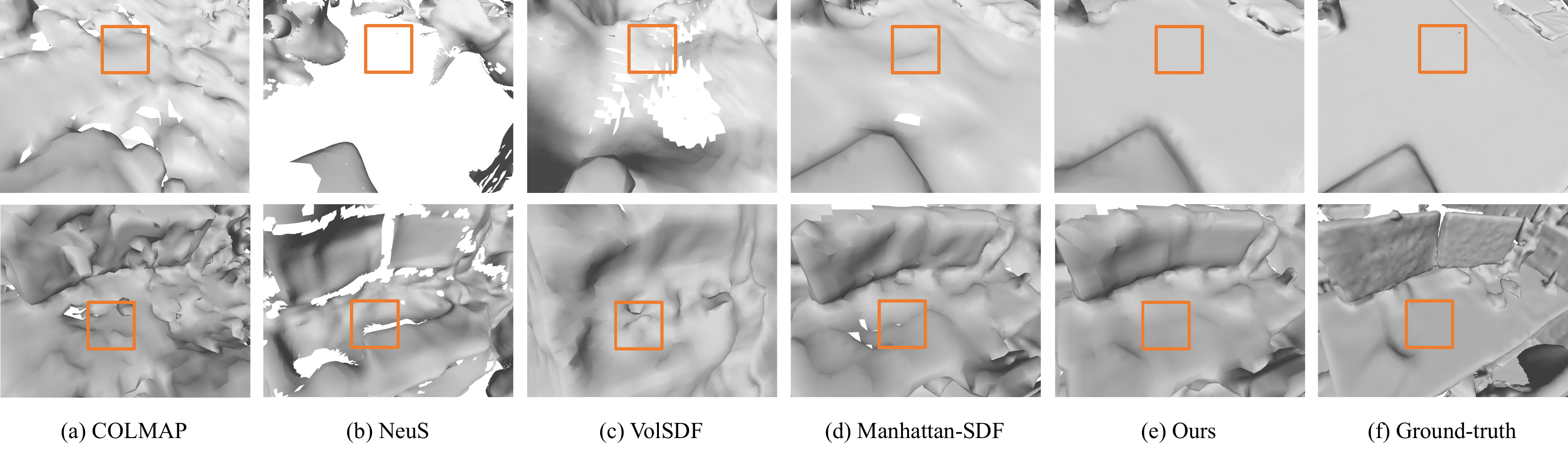}
	\caption{Visualization compared with other methods on the scenes with small visible areas of walls and floors. COLMAP, NeuS, and VolSDF do not lead to complete reconstructions. Manhattan-SDF cannot put effective Manhattan world assumption constraints on these scenes and produce poor reconstructions on some objects, e.g., the table tennis table and computer desk. Instead, our method relies not only on the wall and floor regions but also on these small areas of planes on objects more generally. So we can achieve more smooth and more accurate results with more details.}
	\label{fig:bad_wall}
\end{figure*}

\noindent\textbf{Manhattan scenes:} 
\begin{enumerate}[wide, labelwidth=!, labelindent=0pt]
	\item We compare P$^2$SDF with other methods on the four scenes on the ScanNet dataset \cite{dai2017scannet}, and the four scenes on the 7-Scenes \cite{shotton2013scene} from Manhattan-SDF \cite{guo2022neural}. As can be seen in \cref{tab:ori_scenes}, our method achieves state-of-the-art results on these scenes. As can be seen in \cref{fig:ori_scenes}, COLMAP and NeuS cannot produce proper reconstruction results. VolSDF leads to noisy reconstructions without additional constraints. Manhattan-SDF improves the smooth results in the regions such as floor, wall and bed, but details are missing in some small planar areas such as the sweeping robot. In contrast, our method can get a more detailed surface in these regions while achieving compelling results.
	\item We also conduct a comparison on the other scenes on ScanNet dataset. In \cref{tab:bad_iou}, we select 4 scenes where DeepLab V3+ \cite{chen2018encoder} segmentation network predicts poorly on the wall and floor regions in the ScanNet dataset. The results show that Manhattan-SDF \cite{guo2022neural} degrades on these scenes restricted by the inaccurate predictions of the pre-trained segmentation network. As can be seen in \cref{fig:bad_iou}, COLMAP, NeuS, and VolSDF cannot achieve competitive results. Manhattan-SDF produces unreasonable reconstruction results in the poorly predicted segmentation regions, e.g., the wall region near the paper holder. Compared with these approaches, our approach can produce more details and smooth surfaces, achieving better results. The comparison between our generated planes and the semantic segmentation results used in Manhattan-SDF can be seen in \cref{fig:comparison_seg_sem}. DeepLab V3+ predicts poorly in some scenes, while ours give a better segmentation results.
	
	\begin{table}[htbp]
		\centering
		\small
		\caption{Results on the cluttered Manhattan scenes with small visible areas of walls and floors. We choose four scenes from the ScanNet dataset. Results show that our method can still achieve the best performance in these challenging scenes.}
		\begin{tabular}{l|ccccc}
			\toprule
			\makebox[0.08\linewidth][l]{Method} & \makebox[0.09\linewidth][c]{Acc $\downarrow$} & \makebox[0.09\linewidth][c]{Comp $\downarrow$} & \makebox[0.09\linewidth][c]{Prec $\uparrow$} & \makebox[0.09\linewidth][c]{Recall $\uparrow$} & \makebox[0.09\linewidth][c]{\textbf{F-score} $\uparrow$} \\
			\midrule
			NeRF \cite{mildenhall2021nerf}    & 0.448 & 0.566 & 0.042 & 0.010  & 0.015 \\
			VolSDF \cite{yariv2021volume}     & 0.322 & 0.236 & 0.243 & 0.159 & 0.184 \\
                NeuS \cite{wang2021neus}          & 0.124 & 0.156 & 0.370 & 0.285 & 0.320 \\
			Manhattan-SDF \cite{guo2022neural} & 0.103 & 0.148 & 0.442 & 0.361 & 0.396 \\
			COLMAP \cite{schoenberger2016sfm} & 0.160 & 0.136 & 0.404 & 0.406 & 0.402 \\
			\midrule
			Ours                 & \textbf{0.093} & \textbf{0.132} & \textbf{0.494} & \textbf{0.408} & \textbf{0.445} \\
			\bottomrule
		\end{tabular}
		\label{tab:bad_wall}
	\end{table}
	
	\begin{figure}[htbp]
		\centering
		\includegraphics[width=1.0\linewidth]{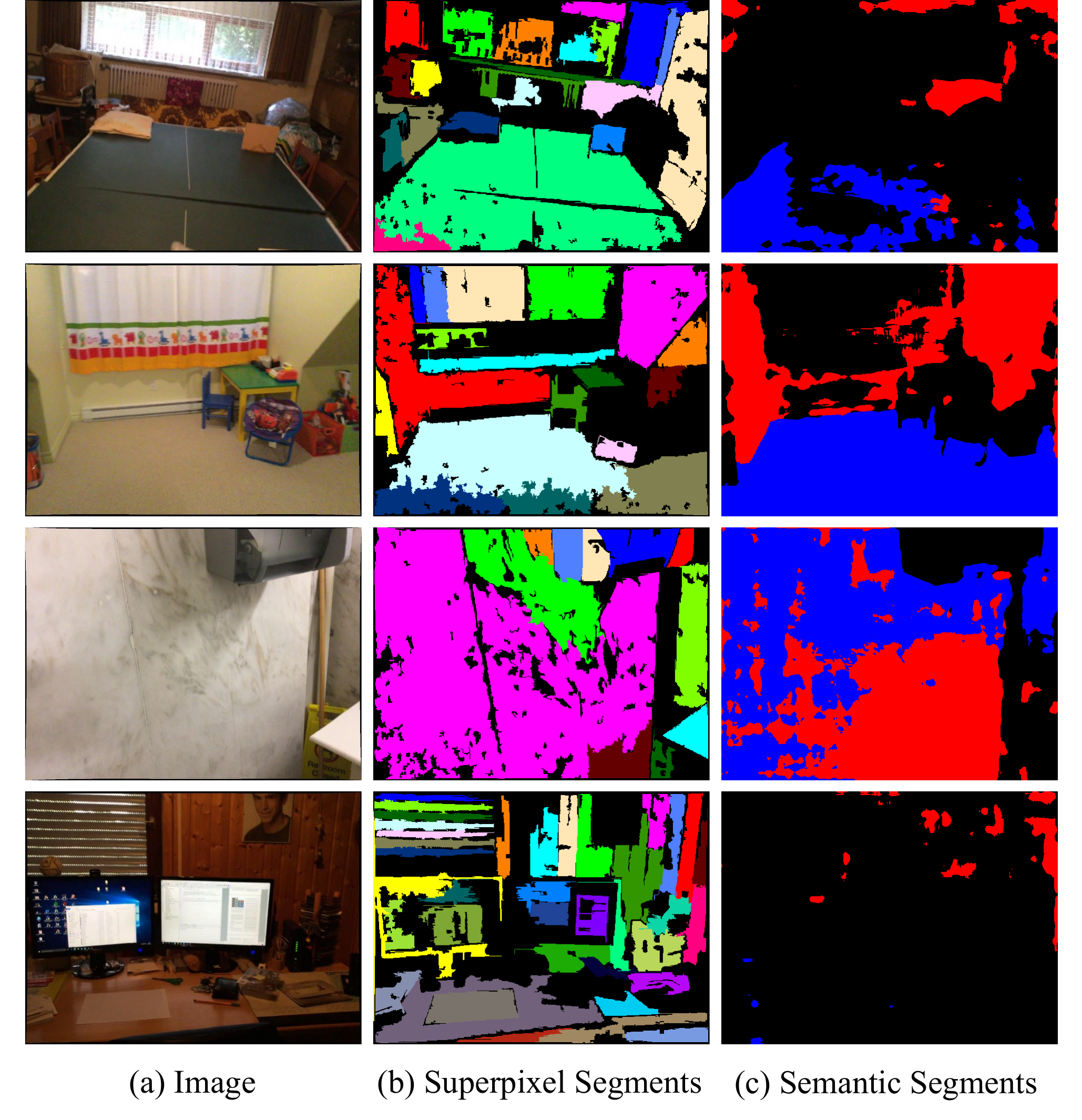}
		\caption{The comparison between plane segments by super-pixel (b) and the semantic segments for wall and floor by DeepLab V3+ (c). The floors are masked with red color, and the walls are masked with blue color in (c). DeepLab V3+ predicts poorly in some scenes. It also predicts nearly nothing in the scenes where walls and floors are heavily occupied. In contrast, super-pixel approach we use gives a better segmentation result in these scenes.}
		\label{fig:comparison_seg_sem}
	\end{figure}
	\item In \cref{tab:bad_wall}, we select the scenes where the walls and floors are heavily occupied by other objects. Manhatan-SDF degrades due to insufficient constraints on the small area of walls and floors, while our proposed plane regularization can still work on the small areas of planar regions on the objects and performs better. In \cref{fig:comparison_seg_sem}, DeepLab V3+ also predicts nearly nothing in the scenes where walls and floors are heavily occupied. In contrast, super-pixel approach performs better in these scenes. Results validate the effectiveness of our proposed P$^2$SDF. The visualization can be in \cref{fig:bad_wall}. COLMAP, NeuS, and VolSDF do not lead to complete reconstructions. Manhattan-SDF cannot put effective Manhattan world assumption constraints on these scenes and produce poor reconstructions on some objects, e.g., the regions on the table tennis table and computer desk. Instead, our method benefits not only on the wall and floor regions but also on these small areas of planes on objects more generally. So we can achieve more smooth and more accurate results with more details.
\end{enumerate} 

\begin{table}[htbp]
	\centering
	\small
	\caption{Results on the non-Manhattan scenes. We choose four non-Manhattan scenes from the ScanNet dataset. Our method surpasses other methods on challenging scenes.}
	\begin{tabular}{l|ccccc}
		\toprule
		\multicolumn{6}{c}{Non-Manhattan} \\
		\toprule
		\makebox[0.08\linewidth][l]{Method} & \makebox[0.09\linewidth][c]{Acc $\downarrow$} & \makebox[0.09\linewidth][c]{Comp $\downarrow$} & \makebox[0.09\linewidth][c]{Prec $\uparrow$} & \makebox[0.09\linewidth][c]{Recall $\uparrow$} & \makebox[0.09\linewidth][c]{\textbf{F-score} $\uparrow$} \\
		\midrule
		NeRF \cite{mildenhall2021nerf}    & 0.538 & 0.760 & 0.036 & 0.011 & 0.016 \\
		NeuS \cite{wang2021neus}          & 0.246 & 0.349 & 0.173 & 0.120 & 0.140 \\
		VolSDF \cite{yariv2021volume}     & 0.419 & 0.223 & 0.217 & 0.199 & 0.204 \\
		COLMAP \cite{schoenberger2016sfm} & 0.134 & 0.195 & 0.440 & 0.358 & 0.389 \\
            Manhattan-SDF \cite{guo2022neural} & 0.109 & 0.132 & 0.516 & 0.450 & 0.478 \\
		\midrule
		Ours                 & \textbf{0.105} & \textbf{0.131} & \textbf{0.542} & \textbf{0.495} & \textbf{0.515} \\
		\bottomrule
	\end{tabular}
	\label{tab:non_manhattan}
\end{table}

\begin{figure*}[htbp]
	\centering
	\includegraphics[width=1.0\linewidth]{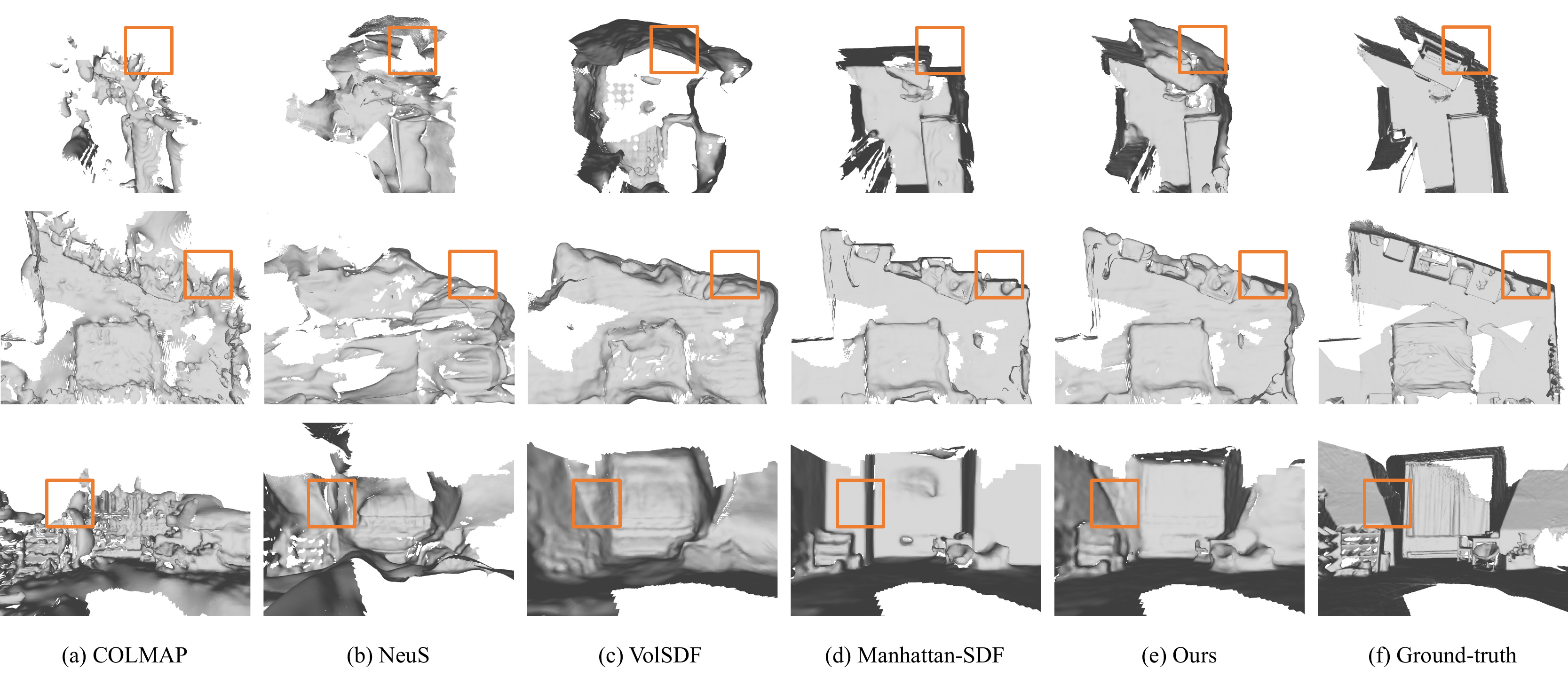}
	\caption{Visualization compared with other methods on the non-Manhattan scenes. COLMAP, NeuS, and VolSDF can only produce rough surface reconstruction results. Manhattan-SDF wrongly predicts the sloping wall as zigzag because of the Manhattan world assumption. Without the Manhattan world assumption, our method achieves reasonable and detailed surfaces, getting better performance and generalizing well.}
	\label{fig:non_manhattan}
\end{figure*}

\begin{table*}[htbp]
	\centering
	\caption{The ablation study of P$^2$SDF. The result is averaged on the four Manhattan scenes from Manhattan-SDF \cite{guo2022neural}. We retrain the VolSDF and add components to it. Results show that every design in our method works and gets the best performance with the combination of all designs.}
	\begin{tabular}{cccc|ccccc}
		\toprule
		Depth & Plane Reg & Weighting & Sampling & Acc $\downarrow$ & Comp $\downarrow$ & Prec $\uparrow$ & Recall $\uparrow$ & \textbf{F-score} $\uparrow$ \\
		\midrule
		&           &            &            & 0.270 & 0.145 & 0.416 & 0.387 & 0.397 \\
		\checkmark &           &            &            & 0.062 & 0.065 & 0.637 & 0.611 & 0.623 \\
		\checkmark &           &            & \checkmark & 0.067 & 0.065 & 0.652 & 0.629 & 0.640 \\
		\checkmark &\checkmark &            &            & 0.059 & 0.067 & 0.680 & 0.636 & 0.657 \\
		\checkmark &\checkmark & \checkmark &            & 0.055 & 0.062 & 0.689 & 0.650 & 0.669 \\
		\checkmark &\checkmark & \checkmark & \checkmark & \textbf{0.055} & \textbf{0.062} & \textbf{0.703} & \textbf{0.664} & \textbf{0.683} \\
		\bottomrule
	\end{tabular}
	\label{tab:ablation}
\end{table*}
\noindent\textbf{Non-Manhattan scenes:} Except Manhattan scenes, there exist lots of non-Manhattan scenes where there are more than three dominant directions. In \cref{tab:non_manhattan}, we choose four non-Manhattan scenes from the ScanNet dataset. The results show that our method performs well on the non-Manhattan scenes and surpasses Manhattan-SDF a lot. The reason is that our method is not constrained the Manhattan world assumption while Manhattan-SDF still imposes the Manhttan world assumption and gets unreasonable result. \cref{fig:non_manhattan} shows that COLMAP, NeuS, and VolSDF can only produce rough surface reconstruction results. Manhattan-SDF wrongly predicts the sloping wall as zigzag because of the Manhattan world assumption. Without the Manhattan world assumption, our method achieves reasonable and detailed surfaces, getting better performance and generalizing well.

\subsection{Ablation Studies}
\subsubsection{The Importance of Different Modules in P$^2$SDF}
\begin{figure*}[htbp]
	\centering
	\includegraphics[width=1.0\linewidth]{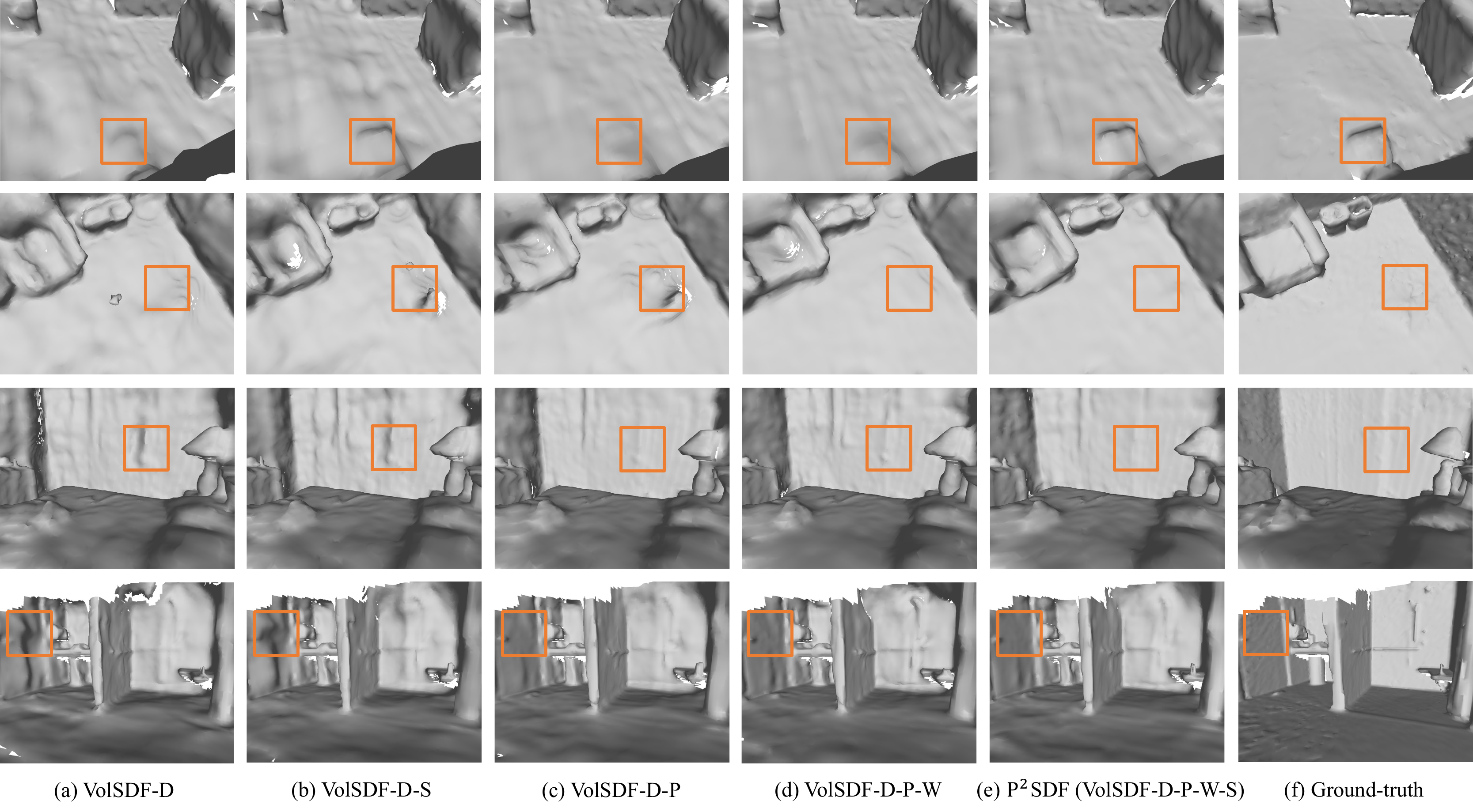}
	\caption{Visualization of the ablation studies of P$^2$SDF on the Manhattan scenes in the ScanNet dataset. Our proposed method can effectively improve reconstruction quality and produce much more detailed and smooth reconstruction results than the baselines. Note that VolSDF-D-S can reconstruct more clear contours of objects such as the sweeping robot and doorframe compared to VolSDF-D, and VolSDF-D-P can reconstruct smoother planes such as the wall region compared to VolSDF-D. VolSDF-D-P-W is also better than VolSDF-D-P. Our P$^2$SDF can maintain the details while also producing smooth and high-quality reconstruction results. (Note: *-D: w/ sparse depth; *-P: w/ our proposed plane regularization; *-W: w/ plane segments fusion based points weighting; *-S: w/ keypoint guided sampling)}
	\label{fig:ablation}
\end{figure*}

\begin{figure*}[htbp]
	\centering
	\includegraphics[width=1.0\linewidth]{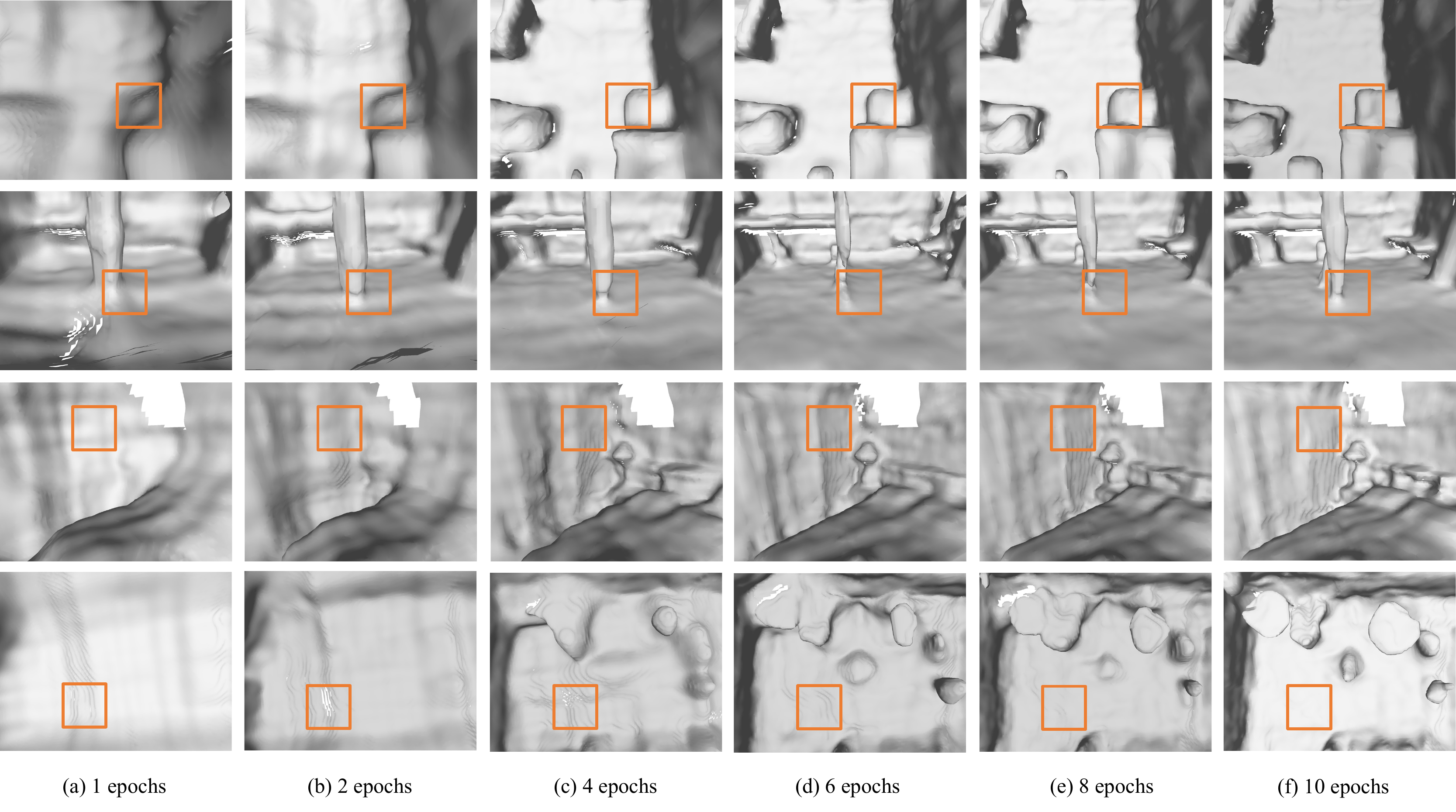}
	\caption{Visualization of the reconstruction results of P$^2$SDF on the Manhattan scenes in the ScanNet dataset at early training epochs. It shows that the contour of the objects such as the boxes and piano become more clear, and some planar areas such as the walls, floor, bed, and table become flatter, gradually. The quality of reconstruction is gradually improving in the training process.}
	\label{fig:ablation_steps}
\end{figure*}

To validate the effectiveness of our proposed method, we conduct ablation studies on the four Manhattan scenes from Manhattan-SDF \cite{guo2022neural}. The results are shown in \cref{tab:ablation}. In \cref{tab:ablation}, the baseline is the original VolSDF \cite{yariv2021volume}. Depth means utilizing the sparse depth from COLMAP \cite{schonberger2016structure} as additional supervision following \cite{roessle2022dense,lin2021efficient,wei2021nerfingmvs,guo2022neural}, which is also adopted in our method. Plane Reg means our proposed plane regularization. Weighting means our proposed plane segments fusion based points weighting. Sampling means the keypoint guided ray sampling strategy. Results show that every design in our proposed P$^2$SDF works, and the combination of all designs gets the best performance. Compared with VolSDF with COLMAP depth as supervision, our proposed plane regularization improves by 0.034 on F-score. F-score further increases by 0.012 by our proposed plane segments fusion based points weighting. The final F-score increases by 0.014 by our proposed keypoint guided ray sampling strategy.  
The visualization can also be seen in \cref{fig:ablation}. The results verify the effectiveness of our method. The visualization of the reconstruction by P$^2$SDF in the training process can be seen in \cref{fig:ablation_steps}. It shows that the regions are becoming planar in the early training.

\begin{table}[htbp]
	\centering
	\caption{The performance of P$^2$SDF with the different number of sample points in rectified plane parameter estimation. The result is the averaged F-score on the four Manhattan scenes from Manhattan-SDF \cite{guo2022neural}. The best performance is achieved with 8192 points. The larger or smaller number of sample points degrades the performance.}
	\begin{tabular}{c|cccccc}
		\toprule
		\makebox[0.2\linewidth][c]{\# of samples} & \makebox[0.02\linewidth][c]{1024} & \makebox[0.02\linewidth][c]{2048} & \makebox[0.02\linewidth][c]{4096} & \makebox[0.02\linewidth][c]{8192} & \makebox[0.02\linewidth][c]{16384} & \makebox[0.02\linewidth][c]{32768} \\
		\midrule
		F-score            & 0.633 & 0.638 & 0.642 & \textbf{0.657} & 0.638 & 0.634 \\
		\bottomrule
	\end{tabular}
	\label{tab:number_samples}
\end{table}
\begin{figure}[htbp]
	\centering
	\includegraphics[width=\linewidth]{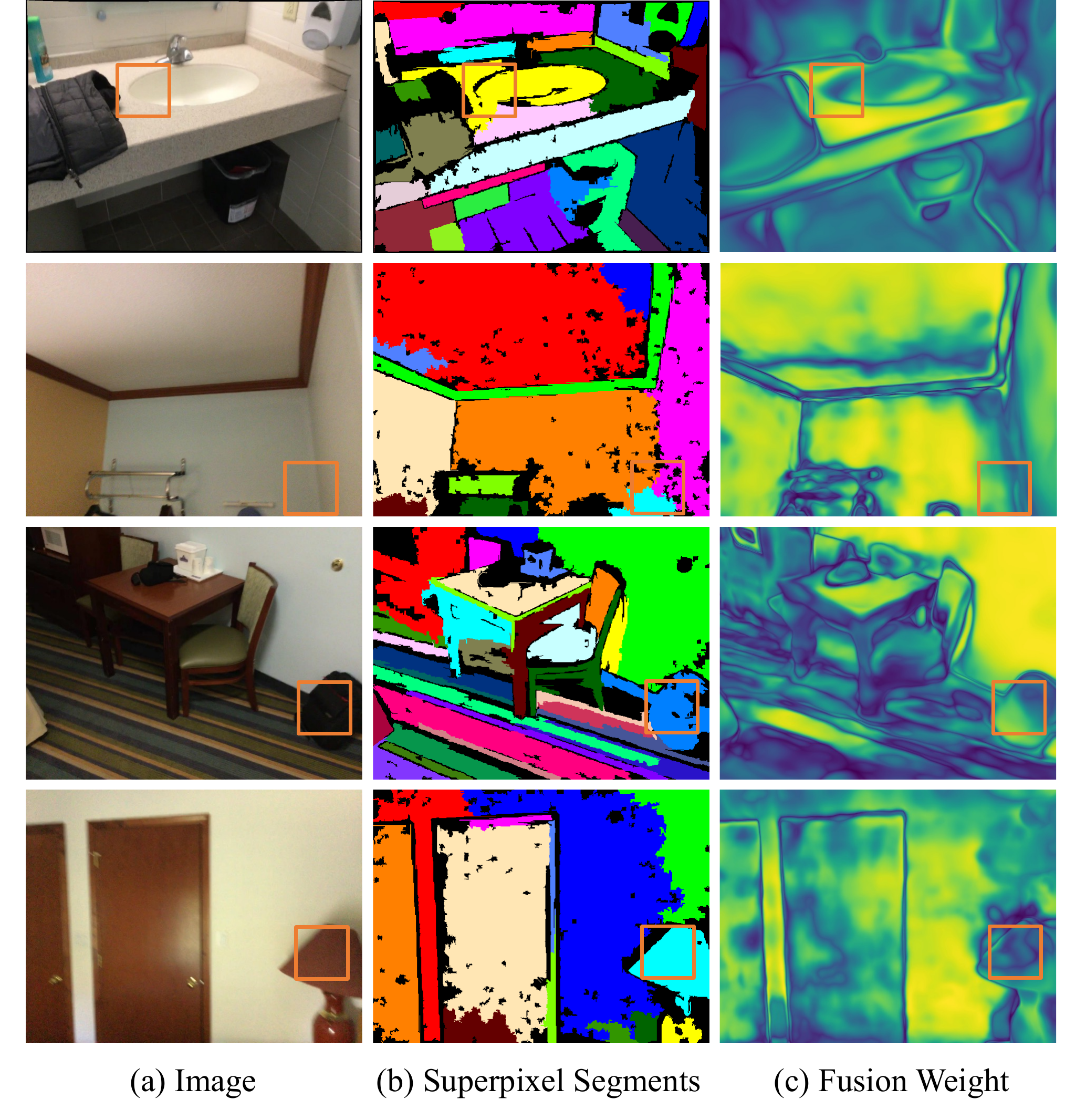}
	\caption{The illustration of the plane segments by super-pixel (b) and the weights of points with plane segments fusion (c). The weights are obtained by plane segments fusion from different views. The weights distribute differently in the whole image. The weights are very small in the mistakenly segmented regions, such as the schoolbag and lamp. In contrast, the weights are large in the accurately segmented regions, such as the ceiling and wall.}
	\label{fig:comparison_seg_weight}
\end{figure}
\subsubsection{Number of Points in Plane Parameter Estimation.} The number of points is important to estimate an accurate plane parameter, therefore, we propose the two-step plane parameter estimation. In \cref{tab:number_samples}, we study the effect of the different number of points in the rectified plane parameter estimation. It can be seen that when the number of sample points is too small, the performance degrades a lot. The reason is that the few sample points make the plane estimation inaccurate. Accurate estimation is in need of sampling sufficient points. When the number of sample points is large, the performance also degrades. The reason may be that the unsupervised plane segments by super-pixel are noisy and inaccurate, therefore it may need some randomness.

\subsubsection{Visualization of Plane Segments Fusion Based Points Weighting}
We visualize the plane segments fusion based points weighting. As can be sen in \cref{fig:comparison_seg_weight}, the weights are small in the mistakenly segmented regions, such as the schoolbag and lamp. In contrast, the weights are large in the accurately segmented regions, such as the ceiling and wall. The weights distribute differently in the whole image, showing the effectiveness of our method.

\subsubsection{Loss Weights} 
\begin{table}[htbp]
	\centering
	\caption{The performance of P$^2$SDF with different loss weights of plane regularization. The result is the averaged F-score on the four Manhattan scenes from Manhattan-SDF \cite{guo2022neural}.}
	\begin{tabular}{c|cccccc}
		\toprule
		\makebox[0.2\linewidth][c]{weight} & \makebox[0.02\linewidth][c]{0.01} & \makebox[0.02\linewidth][c]{0.05} & \makebox[0.02\linewidth][c]{0.1} & \makebox[0.02\linewidth][c]{0.2} & \makebox[0.02\linewidth][c]{0.5} & \makebox[0.02\linewidth][c]{1.0} \\
		\midrule
		F-score            & 0.618 & 0.632 & 0.649 & \textbf{0.657} & 0.640 & 0.592 \\
		\bottomrule
	\end{tabular}
	\label{tab:different_sdf}
\end{table}

\begin{table}[htbp]
	\centering
	\caption{The performance of P$^2$SDF with different loss weights of plane segmentation. The result is the averaged F-score on the four Manhattan scenes from Manhattan-SDF \cite{guo2022neural}.}
	\begin{tabular}{c|ccccccc}
		\toprule
		\makebox[0.2\linewidth][c]{weight} & \makebox[0.02\linewidth][c]{0.001} & \makebox[0.02\linewidth][c]{0.005} & \makebox[0.02\linewidth][c]{0.01} & \makebox[0.02\linewidth][c]{0.05} & \makebox[0.02\linewidth][c]{0.1} \\
		\midrule
		F-score            & 0.649 & 0.650 & \textbf{0.669} & 0.653 & 0.647 \\
		\bottomrule
	\end{tabular}
	\label{tab:different_cross_entropy}
\end{table}
We conduct ablations on different loss weights in P$^2$SDF. Appropriate loss weights in the proposed plane regularization and plane segmentation get the best performance, seen in \cref{tab:different_sdf} and \cref{tab:different_cross_entropy}. Too large loss weights for plane regularization and plane segmentation may hinder the optimization process of neural surface reconstruction, while too small loss weights may lose the effect.

\subsubsection{Keypoint Guided Ray Sampling Strategy}
\begin{table}[htbp]
	\centering
	\caption{The ablation of the keypoint guided ray sampling strategy. The result is the averaged F-score on the four Manhattan scenes from Manhattan-SDF \cite{guo2022neural}.}
	\begin{tabular}{l|cccc}
		\toprule
		\multirow{2}[4]{*}{Method} & \multicolumn{4}{c}{$k$} \\
		\cmidrule{2-5}          & 0.5   & 1     & 1.5   & 2 \\
		\midrule
		Uniform & 0.607 & 0.608 & 0.617 & 0.608 \\
		Gaussian & 0.612 & 0.612 & \textbf{0.640}  & 0.614 \\
		\bottomrule
	\end{tabular}%
	\label{tab:different_saml}
\end{table}
\begin{figure*}[htbp]
	\centering
	\includegraphics[width=1.0\linewidth]{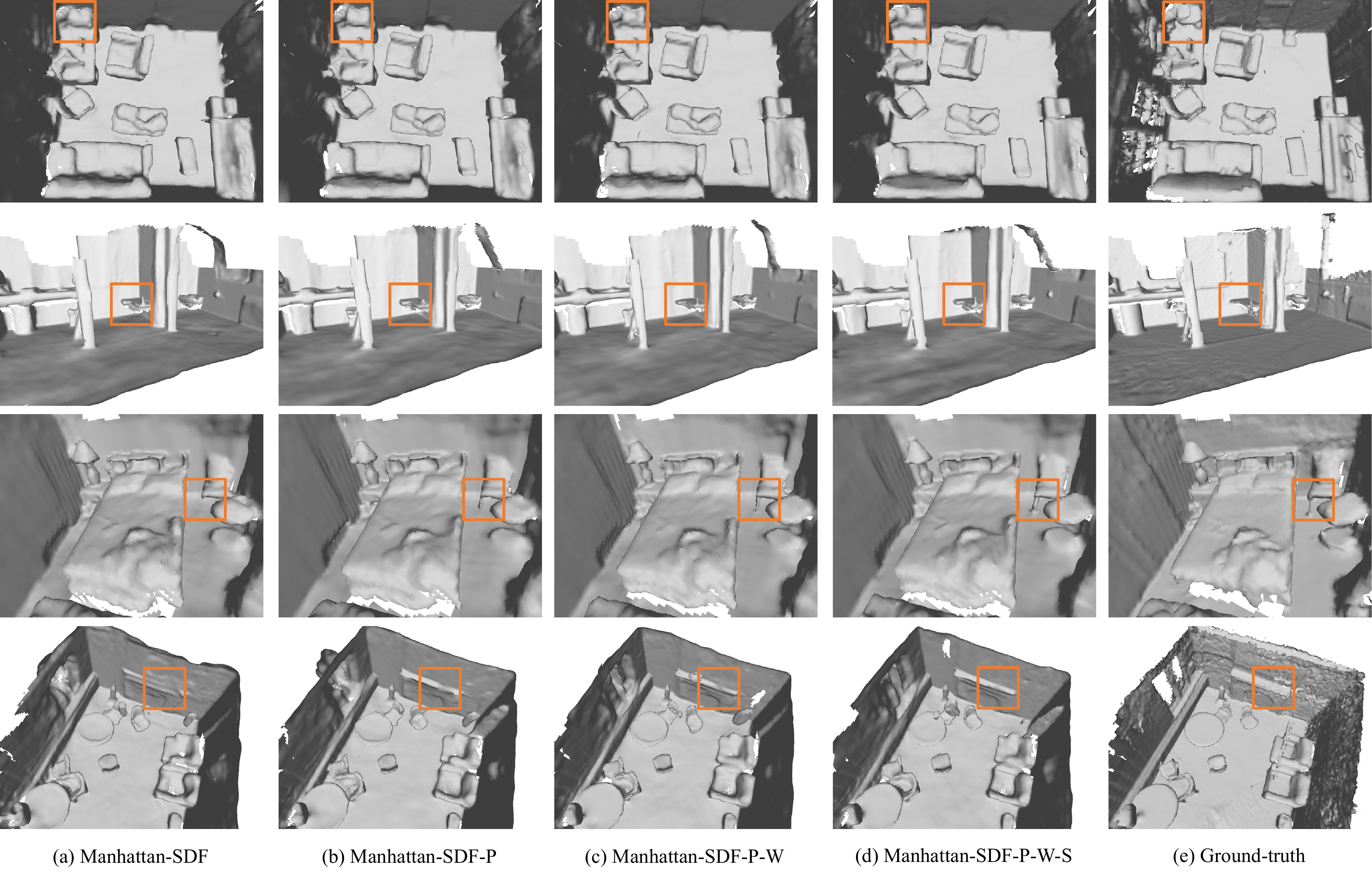}
	\caption{Visualization of the combination with Manhattan-SDF in scenes in the ScanNet dataset. We combine our approach with Manhattan-SDF and further improve the quality and accuracy of the reconstruction. Manhattan-SDF-P can reconstruct smoother planes in the wall and floor regions compared to Manhattan-SDF. Manhattan-SDF-P-W can produce more accurate planes and clearer contours of objects such as the printer and fireplace than Manhattan-SDF-P. Manhattan-SDF-P-W-S can maintain the reconstruction quality of planes and produce much more details results than Manhattan-SDF-P-W, e.g., on the water cup and chair legs. (Note: *-D: w/ sparse depth; *-P: w/ our proposed plane regularization; *-W: w/ plane segments fusion based points weighting; *-S: w/ keypoint guided sampling)}
	\label{fig:combine}
\end{figure*}
We conduct ablation for keypoint guided ray sampling strategy in \cref{tab:different_saml}. We try different sampling weight strategies. Gaussian means that the weights in the patches are set corresponding to the Gaussian distance to the keypoints in \cref{eq:sampl_weight}. Weights are large near the keypoints and small far from them. Uniform means that the weights in the patches centered at keypoints are set uniform and equal to the mean of the corresponding Gaussian weights. The results show that Gaussian performs better than the Uniform weights.

\subsubsection{Combination with Manhattan-SDF}
\begin{table}[htbp]
	\centering
	\caption{Averaged results of the combination of our proposed P$^2$SDF and Manhattan-SDF \cite{guo2022neural} on the four Manhattan scenes from Manhattan-SDF. Results show that our proposed methods are compatible and can get better performance combined with Manhattan-SDF. (Note: *-D: w/ sparse depth; *-P: w/ our proposed plane regularization; *-W: w/ plane segments fusion based points weighting; *-S: w/ keypoint guided sampling)}
	\begin{tabular}{l|ccccc}
		\toprule
		\makebox[0.1\linewidth][l]{Method} & \makebox[0.09\linewidth][c]{Acc $\downarrow$} & \makebox[0.09\linewidth][c]{Comp $\downarrow$} & \makebox[0.09\linewidth][c]{Prec $\uparrow$} & \makebox[0.09\linewidth][c]{Recall $\uparrow$} & \makebox[0.09\linewidth][c]{\textbf{F-score} $\uparrow$} \\
		\midrule
		M-SDF & 0.050  & 0.055 & 0.705 & 0.660 & 0.681 \\
		M-SDF + P & 0.049 & 0.053 & 0.714 & 0.665 & 0.688 \\
		M-SDF + P + W & 0.048 & 0.054 & 0.727 & 0.674 & 0.699 \\
		M-SDF + P + W + S & \textbf{0.046} & \textbf{0.052} & \textbf{0.737} & \textbf{0.683} & \textbf{0.709} \\
		\bottomrule
	\end{tabular}
	\label{tab:combine}
\end{table}
Though our method achieves comparable performance on the four Manhattan scenes  in ScanNet dataset from Manhattan-SDF \cite{guo2022neural} and performs better on other scenes, we combine our method with Manhattan-SDF to achieve better performance. Manhattan-SDF can deal with large areas of planar regions, while our plane regularization can serve as an enhancement on the small areas of plane regions. As can be seen in \cref{tab:combine}, with our proposed plane regularization, the F-score improves by 0.007. Adding plane segments fusion based points weighting, the F-score further improves by 0.011. With our keypoint guided sampling, the F-score further improves by 0.01. The results validate that the effectiveness of our method and our designs are compatible with Manhattan-SDF. The visualization can be seen in \cref{fig:combine}. Manhattan-SDF-P can reconstruct smoother planes in the wall and floor regions compared to Manhattan-SDF. Manhattan-SDF-P-W can produce more accurate planes and clearer contours of objects such as the printer and fireplace than Manhattan-SDF-P. Manhattan-SDF-P-W-S can maintain the reconstruction quality of planes and produce much more details results than Manhattan-SDF-P-W, e.g., on the water cup and chair legs.

\section{Conclusion}
\label{sec:conclusion}
In this work, we present the pseudo plane regularized signed distance field for indoor scene reconstruction. We consider adjacent pixels with similar colors to be on the same pseudo planes and regularize the signed field of the neural implicit surface representation by these planes. To estimate the plane parameters on the fly during training effectively and efficiently, we propose a two-step estimation scheme. The singed distances of the sampled points are regularized to be the distances to the planes. To remedy the noisy and inaccurate plane segments, different weights are assigned to the sampled points in plane estimation as well as the regularization loss. The weights come by fusing the plane segments from different views. The keypoint guided ray sampling strategy is also proposed to avoid redundant sampling in the planar regions and attend to the informative textured regions. The implicit network gets the better reconstruction by this strategy. Experiments show that our method not only achieves competitive reconstruction results in Manhattan scenes but also generalizes well to the non-Manhattan scenes.


%



\ifCLASSOPTIONcompsoc
\section*{Acknowledgments}
\else
\section*{Acknowledgment}
\fi

The authors would like to thank Shenhan Qian for the meaningful discussion and suggestion in this work.

\ifCLASSOPTIONcaptionsoff
\newpage
\fi



%
%
%
\bibliographystyle{IEEEtran}
\bibliography{egbib}
%




\end{document}